%% file: main.tex
\documentclass[journal,10pt,twocolumn]{IEEETran}
\usepackage[utf8]{inputenc}

\usepackage{threeparttable}
\usepackage{cite}
\usepackage{amsmath,amssymb,amsfonts}
\usepackage{algorithmic}
\usepackage{graphicx}
\usepackage{textcomp}
\usepackage{indentfirst}
\usepackage{graphicx}
\usepackage{csquotes}
\usepackage{multirow}
\usepackage{tabularx}
\usepackage{float}
\usepackage{color}
\usepackage{framed}
\usepackage{comment}
\usepackage{caption2}
\usepackage{algorithm2e,setspace}
\usepackage{color, colortbl}
\usepackage{hhline}
\SetKwProg{Fn}{Function}{}{}
\definecolor{Gray}{gray}{0.9}
\newcolumntype{g}{>{\columncolor{Gray}}c}

\usepackage{bbding}

\renewcommand{\figurename}{Figure}
\def\tablename{Table}

\begin{document}

\title{Exploring Adversarial Attack in Spiking Neural Networks with Spike-Compatible Gradient}

\author{Ling Liang, Xing Hu, \IEEEmembership{Member}, \IEEEmembership{IEEE}, Lei Deng, \IEEEmembership{Member}, \IEEEmembership{IEEE}, Yujie Wu, Guoqi Li, \IEEEmembership{Member}, \IEEEmembership{IEEE}, \\Yufei Ding, Peng Li, \IEEEmembership{Fellow}, \IEEEmembership{IEEE}, Yuan Xie, \IEEEmembership{Fellow}, \IEEEmembership{IEEE}\\

\thanks{The work was partially supported by National Science Foundation (Grant No. 1725447), Tsinghua University Initiative Scientific Research Program, Tsinghua-Foshan Innovation Special Fund (TFISF), and National Natural Science Foundation of China (Grant No. 61876215). Ling Liang and Xing Hu contributed equally to this work, corresponding author: Lei Deng. Ling Liang, Xing Hu, Lei Deng, Peng Li, and Yuan Xie are with the Department of Electrical and Computer Engineering, University of California, Santa Barbara, CA 93106, USA (email: \{lingliang, xinghu, leideng, lip, yuanxie\}@ucsb.edu). Yujie Wu and Guoqi Li are with the Department of Precision Instrument, Center for Brain Inspired Computing Research, Tsinghua University, Beijing 100084, China (email: wu-yj16@mails.tsinghua.edu.cn, liguoqi@mail.tsinghua.edu.cn). Yufei Ding is with the Department of Computer Science, University of California, Santa Barbara, CA 93106, USA (email: yufeiding@cs.ucsb.edu).} } 

\maketitle

\begin{abstract}

Spiking neural network (SNN) is broadly deployed in neuromorphic devices to emulate the brain function. In this context, SNN security becomes important while lacking in-depth investigation, unlike the hot wave in deep learning. To this end, we target the adversarial attack against SNNs and identify several challenges distinct from the ANN attack: i) current adversarial attack is based on gradient information that presents in a spatio-temporal pattern in SNNs, hard to obtain with conventional learning algorithms; ii) the continuous gradient of the input is incompatible with the binary spiking input during gradient accumulation, hindering the generation of spike-based adversarial examples; iii) the input gradient can be all-zeros (i.e. vanishing) sometimes due to the zero-dominant derivative of the firing function, prone to interrupt the example update. 

Recently, backpropagation through time (BPTT)-inspired learning algorithms are widely introduced into SNNs to improve the performance, which brings the possibility to attack the models accurately given spatio-temporal gradient maps. We propose two approaches to address the above challenges of gradient-input incompatibility and gradient vanishing. Specifically, we design a gradient-to-spike (G2S) converter to convert continuous gradients to ternary ones compatible with spike inputs. Then, we design a restricted spike flipper (RSF) to construct ternary gradients that can randomly flip the spike inputs with a controllable turnover rate, when meeting all-zero gradients. Putting these methods together, we build an adversarial attack methodology for SNNs trained by supervised algorithms. Moreover, we analyze the influence of the training loss function and the firing threshold of the penultimate layer, which indicates a ``trap'' region under the cross-entropy loss that can be escaped by threshold tuning. Extensive experiments are conducted to validate the effectiveness of our solution, showing 99\%+ attack success rate on most benchmarks, which is the best result in SNN attack. Besides the quantitative analysis of the influence factors, we evidence that SNNs are more robust against adversarial attack than ANNs. This work can help reveal what happens in SNN attack and might stimulate more research on the security of SNN models and neuromorphic devices.
\end{abstract}


\input{text/Introduction.tex}

\input{text/Background.tex}

\input{text/Approach.tex}
\input{text/Results.tex}

\input{text/Conclusion.tex}

\bibliography{./ref/ref}

\input{text/biography.tex}
\input{text/Appendix}

\end{document}

%% file: text/Introduction.tex
\section{Introduction}\label{sec:Intro}

Spiking neural networks (SNNs) \cite{maass1997networks} closely mimic the behaviors of neural circuits via spatio-temporal neuronal dynamics and event-drive activities (1-spike or 0-nothing). They have shown promising ability in processing dynamic and noisy information with high efficiency \cite{deng2020rethinking,maass2014noise} and have been applied in a broad spectrum of tasks such as optical flow estimation \cite{Haessig2017Spiking}, spike pattern recognition \cite{wu2019direct}, SLAM \cite{Vidal2018Ultimate}, probabilistic inference \cite{maass2014noise}, heuristically solving NP-hard problem \cite{jonke2016solving}, quickly solving optimization problem \cite{davies2018loihi}, sparse representation \cite{shi2017object}, robotics \cite{hwu2017self}, and so forth. Besides the algorithm research, SNNs are widely deployed in neuromorphic devices for low-power brain-inspired computing  \cite{merolla2014million,furber2014spinnaker,davies2018loihi,pei2019towards}.  

With more attention on SNNs from both academia and industry, the security problem becomes quite important. Here we focus on adversarial attack \cite{szegedy2013intriguing}, one of the most popular threat models for neural network security. In adversarial attack, the attacker introduces imperceptible malicious perturbation into the input data, i.e. generating adversarial examples,  to manipulate the model to cross the decision boundary thus misleading the classification result. Usually, there are two categories of approach to realize adversarial attack: content-based and gradient-based. The former directly modifies the semantic information (e.g. brightness, rotation, etc.) of inputs or injects predefined Trojan into inputs \cite{brown2017adversarial, eykholt2017robust, liu2017trojaning, pei2017deepxplore, brendel2017decision}; while the latter modifies inputs according to the input gradient under specified labels \cite{goodfellow2014explaining, kurakin2016adversarial, moosavi2016deepfool, papernot2016limitations, carlini2017towards}. The gradient-based adversarial attack is able to achieve a better attack effectiveness, which is the focus of this work. 

Although adversarial attack is a very hot topic in artificial neural networks (ANNs), it is rarely studied in the SNN domain. We identify several challenges in attacking an SNN model using adversarial examples. First, the input gradient in SNNs presents as a spatio-temporal pattern that is hard to obtain with traditional learning algorithms like the gradient-free unsupervised learning \cite{diehl2015unsupervised, kheradpisheh2018stdp} and spatial-gradient-only ANN-to-SNN-conversion learning \cite{diehl2015fast}. Second, the gradients are continuous values, incompatible with the binary spiking inputs. This data format incompatibility impedes the generation of spike-based adversarial examples via gradient accumulation. At last, there is severe gradient vanishing when the gradient crosses the step firing function with a zero-dominant derivative, which will interrupt the update of adversarial examples. In fact, there are several prior studies on SNN attack using trial-and-error input perturbation or transferring the techniques proposed for ANN attack. Specifically, the input can be perturbed in a trial-and-error manner by simply monitoring the output change without calculating the gradient \cite{marchisio2019snn,bagheri2018adversarial}; the adversarial examples generated by the substitute ANN counterpart can be inherited to attack the SNN model \cite{sharmin2019comprehensive}. However, they just circumvent rather than directly solve the SNN attack problem, which leads to some drawbacks that will eventually lower down the attack effectiveness. For example, the trial-and-error input perturbation method faces a large search space without the guidance of supervised gradients; the SNN/ANN model conversion method needs extra model transformation and ignores the gradient information in the temporal dimension. 

Recently, the backpropagation through time (BPTT)-inspired supervised learning algorithms \cite{wu2018spatio,jin2018hybrid,bellec2018long,wu2019direct,deng2020rethinking,gu2019stca,neftci2019surrogate} are widely introduced into SNNs for performance boost, which enables the direct acquisition of gradient information in both spatial and temporal dimensions, i.e. spatio-temporal gradient map. This brings opportunity to realize an accurate SNN attack based on spatio-temporal input gradients directly calculated in SNNs without model conversion. Then, to address the mentioned issues of gradient-input incompatibility and gradient vanishing, we propose two approaches. We design a gradient-to-spike (G2S) converter to convert continuous gradients to ternary ones that are compatible with spike inputs. G2S exploits smart techniques including probabilistic sampling, sign extraction, and overflow-aware transformation, which can simultaneously maintain the spike format and control the perturbation magnitude. Then we design a restricted spike flipper (RSF) to construct ternary gradients that can randomly flip the spike inputs when facing all-zero gradient maps, where the turnover rate of inputs is controllable. Under this attack methodology for both untargeted and targeted attacks, we analyze the impact of two important factors on the attack effectiveness: the format of training loss function and the firing threshold. We find a ``trap'' region for the model trained by cross-entropy (CE) loss, which makes it harder to attack when compared to the one trained by mean square error (MSE) loss. Fortunately, the ``trap'' region can be escaped by adjusting the firing threshold of the penultimate layer. We extensively validate our SNN attack methodology on both neuromorphic datasets (e.g. N-MNIST \cite{orchard2015converting} and CIFAR10-DVS \cite{li2017cifar10}) and image datasets (e.g. MNIST \cite{lecun1998gradient} and CIFAR10 \cite{krizhevsky2009learning}), and achieve superior attack results. We summarize our contributions as below:
\begin{itemize}
    \item We identify the challenges of adversarial attack against SNN models, which are quite different from the ANN attack. Then, we realize accurate SNN attack for the first time via spike-compatible spatio-temporal gradient. This work can help reveal what happens in attacking SNNs and might stimulate more research on the security of SNN models and neuromorphic devices.
    
    \item We design a gradient-to-spike (G2S) converter to address the gradient-input incompatibility problem and a restricted spike flipper (RSF) to address the gradient vanishing problem, which form a gradient-based adversarial attack methodology against SNNs trained by supervised algorithms. The perturbation magnitude is well controlled in our design.
    
    \item We explore the influence of the training loss function and the firing threshold of the penultimate layer, and propose threshold tuning to improve the attack effectiveness.
    
    \item Extensive experiments are conducted on both neuromorphic and image datasets, where our methodology shows 99\%+ attack success rate in most cases, which is the best result on SNN attack. Besides, we demonstrate the higher robustness of SNNs against adversarial attack when compared with ANNs. 
\end{itemize}

The rest of this paper is organized as follows: Section \ref{sec:Preliminary} provides some preliminaries of SNNs and adversarial attack; Section \ref{sec:Approach} discusses the challenges in SNN attack and our differences with prior work; Section \ref{sec:attack} and Section \ref{sec: knobs} illustrate our attack methodology and the two factors that can affect the attack effectiveness; The experimental setup and the result analyses are shown in Section \ref{sec:Results}; Finally, Section \ref{sec:Conclusion} concludes and discusses the paper.

%% file: text/Background.tex
\section{Preliminaries}\label{sec:Preliminary}

\subsection{Spiking Neural Networks}\label{sec:snn}

\begin{figure}[!htbp]
\centering
\includegraphics[width=0.38\textwidth]{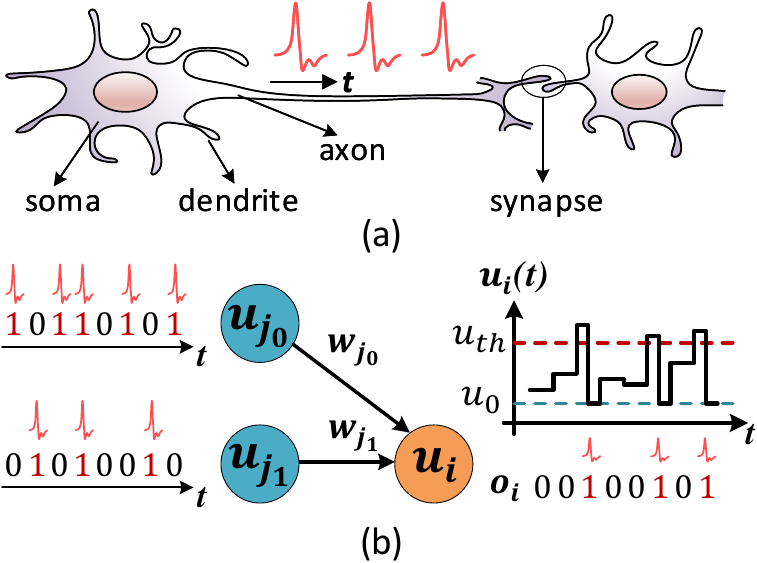}
\caption{Introduction of SNNs: (a) neuronal components; (b) computing model.}
\label{fig:snn}
\end{figure}

Inspired by the biological neural circuits, SNN is designed to mimic their behaviors. A spiking neuron is the basic structural unit, as shown in Figure \ref{fig:snn}, which is comprised of dendrite, soma and axon; many spiking neurons connected by weighted synapses form an SNN, in which the binary spike events carry information for inter-neuron communication. Dendrite integrates the weighted pre-synaptic inputs, and soma consequently updates the membrane potential and determines whether to fire a spike or not. When the membrane potential crosses a threshold, a spike will be fired and sent to post-neurons through axon.  

 The leaky integrate-and-fire (LIF) model \cite{gerstner2014neuronal} is the most widely adopted SNN model. The behavior of each LIF neuron can be briefly expressed as 
\begin{equation}
    \label{equ:lif}
    \begin{cases}
\tau \frac{du(t)}{dt} = -u(t) + \sum_j w_j o_j(t)\\
\begin{cases}
o(t)=1~\&~u(t)=u_0,~\text{if}~u(t)\geq u_{th}\\
o(t)=0,~\text{if}~u(t)< u_{th}~
\end{cases}
\end{cases}
\end{equation} 
where $t$ denotes the time step, $\tau$ is a time constant, and $u$ and $o$ represent the membrane potential and resulting output spike, respectively. $w_j$ is the synaptic weight between the $j$-th pre-neuron and the current neuron, and $o_j$ is the output spike of the $j$-th pre-neuron (also as the input spike of the current neuron). $u_{th}$ is the mentioned firing threshold and $u_0$ is the reset potential used after firing a spike. Note that a spike should be modeled as the Dirac delta function in the continuous time domain; otherwise, it cannot increase the potential.

The network structure of feedforward SNNs can be similar with that of ANNs, including convolutional (Conv) layer, pooling layer, and fully-connected (FC) layer. The network inputs can be spike events captured by dynamic vision sensors \cite{Lichtsteiner2008A} (i.e. neuromorphic datasets) or converted from normal image datasets through Bernoulli sampling \cite{deng2020rethinking}. The classification is conducted based on the spikes of the output layer.

\subsection{Gradient-based Adversarial Attack}\label{sec:adversarial_attack}

We take the gradient-based adversarial attack in ANNs as an illustrative example. The neural network is actually a map from inputs to outputs, i.e. $y=f(x)$, where $x$ and $y$ denote inputs and outputs, respectively, and $f: R^m \to R^n$ is the map function. Usually, the inputs are static images in convolutional neural networks (CNNs). In adversarial attack, the attacker attempts to manipulate the victim model to produce incorrect outputs by adding imperceptible  perturbations $\delta$ in the input images. We define $x'=x+\delta$ as an adversarial example. The perturbation is constrained by $\lVert \delta \rVert_p=\lVert x' - x \rVert_p \leq \epsilon$, where  $\lVert \cdot \rVert_p$ denotes the $p$-norm and $\epsilon$ reflects the maximum tolerable perturbation.

Generally, the adversarial attack can be categorized into untargeted attack and targeted attack according to the different attack goals. Untargeted attack fools the model to classify the adversarial example into any other class except for the original correct one, which can be illustrated as $f(x+\delta)\neq f(x)$. In contrast, for targeted attack, the adversarial example must be classified in to a specified class, i.e. $f(x+\delta)= y_{target}$. With these preliminary knowledge, the adversarial attack can be formulated as an optimization problem as below to search the smallest perturbation:
\begin{equation}
	\begin{cases}
    \underset{\delta}{\text{arg min}} \lVert \delta \rVert_p,~s.t.~f(x+\delta) \neq f(x), ~~\text{if untargeted}\\
    \underset{\delta}{\text{arg min}} \lVert \delta \rVert_p,~s.t.~f(x+\delta)=y_{target},~~\text{if targeted}\\    
\end{cases}.
\end{equation}

There are several widely-adopted adversarial attack algorithms to find an approximated solution of the above optimization problem. Here we introduce two of them: the fast gradient sign method (FGSM) \cite{goodfellow2014explaining} and the basic iterative method (BIM) \cite{kurakin2016adversarial}.

\noindent\textbf{FGSM}. The main idea of FGSM is to generate the adversarial examples based on the gradient information of the input. Specifically, it calculates the gradient map of an input image, and then adds or subtracts the $sign$ of this input gradient map in the original image with multiplying a small scaling factor. The generation of adversarial examples can be formulated as
\begin{equation} \label{equ:fgsm}
	\begin{cases}
	x' = x + \eta \cdot sign(\triangledown_xL(\theta,x,y_{orginal})), ~\text{if untargeted}\\
	x' = x - \eta \cdot sign(\triangledown_xL(\theta,x,y_{target})),~\text{if targeted}
	\end{cases}
\end{equation}
where $L$ and $\theta$ denote the loss function and parameters of the victim model. $\eta$ is used to control the magnitude of the perturbation, which is usually small. In untargeted attack, the adversarial example will drive the output away from the original correct class, which results from the gradient ascent-based input modification; while in targeted attack, the output under the adversarial example goes towards the targeted class, owing to the gradient descent-based input modification.

\noindent\textbf{BIM}. BIM algorithm is actually the iterative version of the above FGSM, which updates the adversarial examples in an iterative manner until the attack succeeds. The generation of adversarial examples in BIM is governed by
\begin{equation}\label{equ:bim}
    \begin{cases}
        x'_{k+1} = x'_k + \eta \cdot sign(\triangledown_{x'_k}L(\theta,x'_k,y_{orginal})), ~\text{if untargeted}\\
        x'_{k+1} = x'_k - \eta \cdot sign(\triangledown_{x'_k}L(\theta,x'_k,y_{target})), ~\text{if targeted}
    \end{cases}
\end{equation}
where $k$ is the iteration index. Specifically, $x'_k$ equals the original input when $k=0$, i.e. $x'_0$. 

In ANNs, several advanced attack methods can be potentially extended beyond BIM based algorithm by optimizing the perturbation bound~\cite{moosavi2016deepfool, papernot2016limitations, carlini2017towards, brendel2017decision} or avoiding the gradient calculation \cite{brown2017adversarial, eykholt2017robust, liu2017trojaning, pei2017deepxplore}. In this work, we aim at the preliminary exploration of an effective gradient-based SNN attack, thus adopting the most classic BIM algorithm in our design. We leave the SNN attack with different approaches in future work.

%% file: text/Approach.tex
\begin{figure*}[!htbp]
\centering
\includegraphics[width=0.9\textwidth]{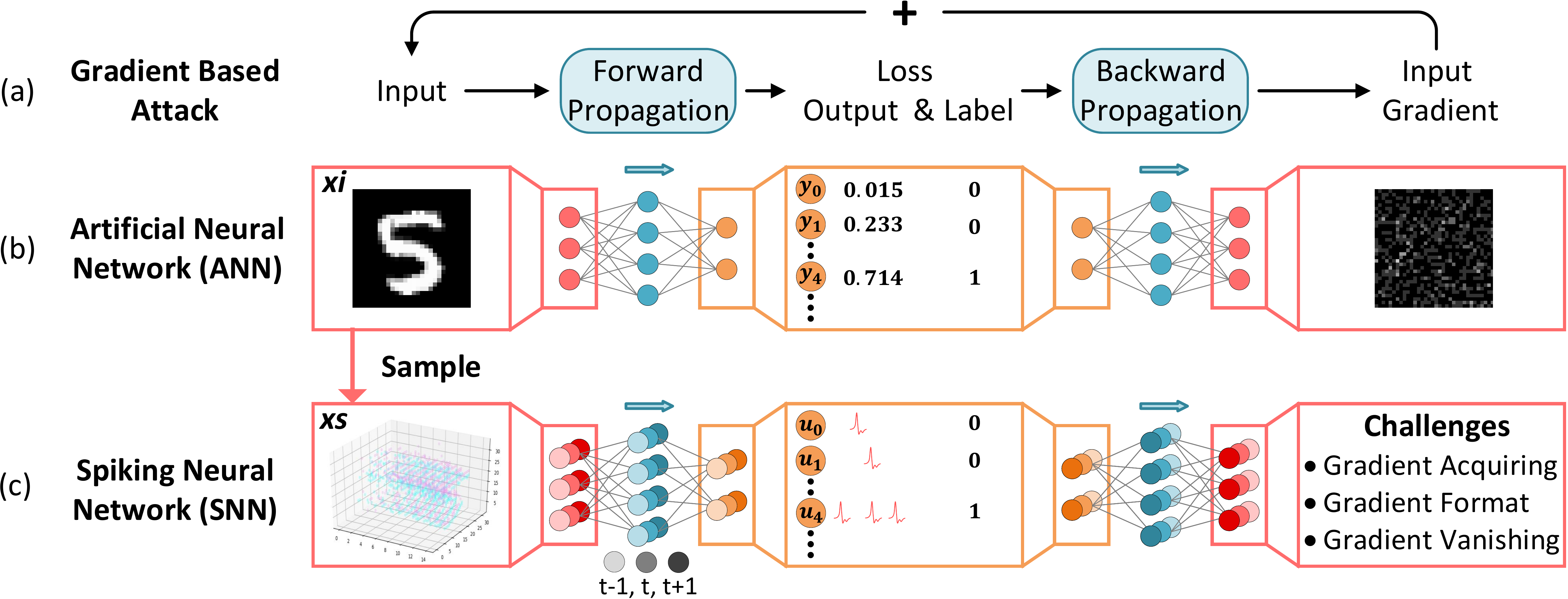}
\caption{Illustration of gradient-based adversarial attack: (a) overall flow including forward pass, backward pass, and input update; (b) adversarial attack in ANNs; (c) adversarial attack in SNNs and its challenges. $xi$ and $xs$ represent an input in image and spike formats, respectively.}
\label{fig:attack_ann_snn}
\end{figure*}

\section{Challenges in SNN Attack}\label{sec:Approach}

Even though the attack methodology can be independent of how the model is trained (e.g., gradient-free unsupervised learning \cite{tan2019exploring} or gradient-based supervised learning \cite{sharmin2019comprehensive}) and it is not necessary to compute gradients when finding adversarial examples (e.g., using trial-and-error methods \cite{marchisio2019snn,bagheri2018adversarial}), we take SNN models trained by BPTT with high recognition accuracy for example and focus on the spatio-temporal gradient based attack due to the potential for high attack success rate. Therefore, all our following discussions about the challenges are restricted in this context. Figure \ref{fig:attack_ann_snn}(a) briefly illustrates the work flow of adversarial attack based on gradients. There are three stages: forward pass to obtain the model prediction, backward pass to calculate the input gradient, and input update to generate the adversarial example. This flow is straightforward to implement in ANNs, as shown in Figure \ref{fig:attack_ann_snn}(b), which is very similar to the ANN training. The only difference lies in the input update that replaces the parameter update in a normal ANN training. However, the case becomes complicated in the SNN scenario, where the processing is based on binary spikes with temporal dynamics rather than continuous activations with immediate response. According to Figure \ref{fig:attack_ann_snn}(c), we attempt to identify the challenges in SNN attack to distinguish from the ANN attack and compare our solution with prior studies in the following subsections.

\subsection{Challenges and Solutions}

\noindent\textbf{Acquiring Spatio-temporal Gradients.} In feedforward ANNs, both the activations and gradients involve only the spatial dimension without temporal components. For each feature map, its gradient during the backward propagation is still in a 2D shape. Whereas, each gradient map becomes 3D in SNNs due to the additional temporal dimension. It is difficult to acquire the spatio-temporal gradients using conventional SNN learning algorithms for the generation of adversarial examples with both spatial and temporal components. For example, the unsupervised learning rules such as spike timing dependent plasticity (STDP) \cite{diehl2015unsupervised} update synapses according to the activities of local neurons, which cannot help calculate the input gradients. The ANN-to-SNN-conversion learning methods \cite{diehl2015fast} simply convert an SNN learning problem into an ANN one with only spatial information, leading to the incapability in capturing temporal input gradients. Recently, the backpropagation through time (BPTT)-inspired learning algorithm \cite{wu2018spatio,jin2018hybrid,bellec2018long,wu2019direct,deng2020rethinking,gu2019stca} is broadly studied to improve the accuracy of SNNs. This emerging supervised learning promises accurate SNN attack via the direct acquisition of input gradients in both spatial and temporal dimensions, which is adopted by us.

\noindent\textbf{Incompatible Format between Gradients and Inputs.} The input gradients are in continuous values, while the SNN inputs are in binary spikes (see the left of Figure \ref{fig:attack_ann_snn}(c), each point represents a spike event, i.e. ``1''; otherwise it is ``0''). This data format incompatibility impedes the generation of spike-based adversarial examples if we consider the conventional gradient accumulation. In this work, we propose a gradient-to-spike (G2S) converter to convert continuous gradients to spike-compatible ternary gradients. This design exploits probabilistic sampling, sign extraction, and overflow-aware transformation, which can simultaneously maintain the spike format and control the perturbation magnitude. 

\noindent\textbf{Gradient Vanishing Problem.} The thresholded spike firing of the LIF neuron, as mentioned in Equation (\ref{equ:lif}), is actually a step function that is non-differentiable. To address this issue, a Dirac-like function is introduced to approximate the derivative of the firing activity \cite{wu2018spatio}. However, this approximation brings abundant zero gradients outside the gradient window (to be shown latter), leading to severe gradient vanishing during backpropagation. We find that the input gradient map can be all-zero sometimes, which interrupts the gradient-based update of adversarial examples. To this end, we propose a restricted spike flipper (RSF) to construct ternary gradients that can randomly flip the binary inputs in the case of all-zero gradients. We use a baseline sampling factor to bound the overall turnover rate, making the perturbation magnitude controllable. 

\subsection{Comparison with Prior Work on SNN Attack}\label{sec:cha:comparison}

The study on SNN attack is rarely seen, which is still in its infant stage. We only find several related works talking about this topic. In this subsection, we summarize their approaches and clarify our differences compared with them.

\begin{table*}[!htbp]
\caption{Comparison with prior work on SNN attack.}
\label{tab:related_work}
\vspace{3pt}
\centering
\renewcommand\arraystretch{1.3}
\resizebox{0.7\textwidth}{!}{
\begin{tabular}{c | c | c | c | c }
\hline \hline
\textbf{Attack Method} & \textbf{Data Source} & \textbf{Spatio-temporal Gradient} & \textbf{Computational Complexity} & \textbf{Attack Effectiveness} \\
\hline
Trial-and-Error \cite{marchisio2019snn} & Image & \XSolidBrush & $Iter \times N \times 2C_{FP}$ & Low \\
Trial-and-Error \cite{bagheri2018adversarial} & Spike & \XSolidBrush & $Iter \times N \times C_{FP}$ & Low \\
Model Conversion \cite{sharmin2019comprehensive} & Image & \XSolidBrush & $Iter \times (C_{FP} + C_{BP})$ & Low\\
\textbf{This Work} & Spike/Image & \Checkmark & $Iter\times (C_{FP} + C_{BP})$ & High\\
\hline \hline
\end{tabular}}
\end{table*}

\noindent \textbf{Trial-and-Error Input Perturbation.} Such attack algorithms perturb inputs in a trial-and-error manner by monitoring the variation of outputs. For example, A. Marchisio et al. \cite{marchisio2019snn} modify the original image inputs before spike sampling. They first select a block of pixels in the images, and then add a positive or negative unit perturbation onto each pixel. During this process, they always monitor the output change to determine the perturbation until the attack succeeds or the perturbation exceeds a threshold. However, this image-based perturbation is not suitable for the data sources with only spike events \cite{orchard2015converting,li2017cifar10}. In contrast, A. Bagheri et al. \cite{bagheri2018adversarial} directly perturb the spike inputs rather than the original image inputs. The main idea is to flip the input spikes and also monitor the outputs. 

\noindent \textbf{SNN/ANN Model Conversion.} S. Sharmin et al. \cite{sharmin2019comprehensive} convert the SNN attack problem into an ANN one. They first build an ANN substitute model that has the same network structure and parameters copied from the trained SNN model. The gradient-based adversarial attack is then conducted on the built ANN counterpart to generate the adversarial examples. 

These existing works suffer from several drawbacks that would eventually degrade the attack effectiveness. For the trial-and-error input perturbation methods, the computational complexity is quite high due to the large search space without the guidance of supervised gradients. Specifically, each selected element of the inputs needs to run the forward pass once (for spike perturbation) or twice (for image perturbation) to monitor the outputs. The total computational complexity is $Iter\times N\times C_{FP}$, where $Iter$ is the number of attack iterations, $N$ represents the size of search space, and $C_{FP}$ is the computational cost of each forward pass. This complexity is much higher than the normal one, i.e. $Iter\times (C_{FP}+C_{BP})$, due to the large $N$. Because it is difficult to find the optimal perturbation in such a huge space, the attack effectiveness cannot be satisfactory given a limited search time in reality. Regarding the SNN/ANN model conversion method, an extra model transformation is needed and the temporal information is lost during the SNN-to-ANN conversion. Using a different model to find gradients and the missing of temporal components will compromise the attack effectiveness in the end. Moreover, this method is not applicable to the image-free spiking data sources without the help of extra signal conversion.

Compared with the above works we calculate the gradients in both spatial and temporal dimensions without extra model conversion, which matches the natural SNN behaviors. As a result, the spatio-temporal input gradients can be acquired in a supervised manner, laying a foundation for effective attack. Then, the proposed G2S and RSF enable the generation of spiking adversarial examples based on the continuous gradients even if when meeting the gradient vanishing. This direct generation of spiking adversarial examples makes our methodology suitable for the image-free spiking data sources. For the SNN models using image-based data sources, our solution is also applicable with a simple temporal aggregation of spatio-temporal gradients. In summary, Table \ref{tab:related_work} shows the differences between our work and prior work. The attack effectiveness in the table is a joint indicator of the computational complexity, compatibility of input data formats, and attack success rate (under the same restriction of perturbation bound). If a method can achieve high attack success rate with relative low complexity and better compatibility of input data formats, we think it has high effectiveness.

Please note that we focus on the white-box attack in this paper. Specifically, in the white-box attack scenario, the adversary knows the network structure and model parameters (e.g. weights, $u_{th}$, etc.) of the victim model. The reason of this scenario selection lies in that the white-box attack is the fundamental step to understand adversarial attack, which is more appropriate for the preliminary exploration on the direct adversarial attack against SNNs based on accurate gradients. Furthermore, the methodology built for the white-box attack can be easily transferred to the black-box attack in the future.

\section{Adversarial Attacks against SNNs}\label{sec:attack}

In this section, we first introduce the input data format briefly, and then explain the flow, approach, and algorithm of our attack methodology in detail. 

\begin{figure}[!htbp]
\centering
\includegraphics[width=0.48\textwidth]{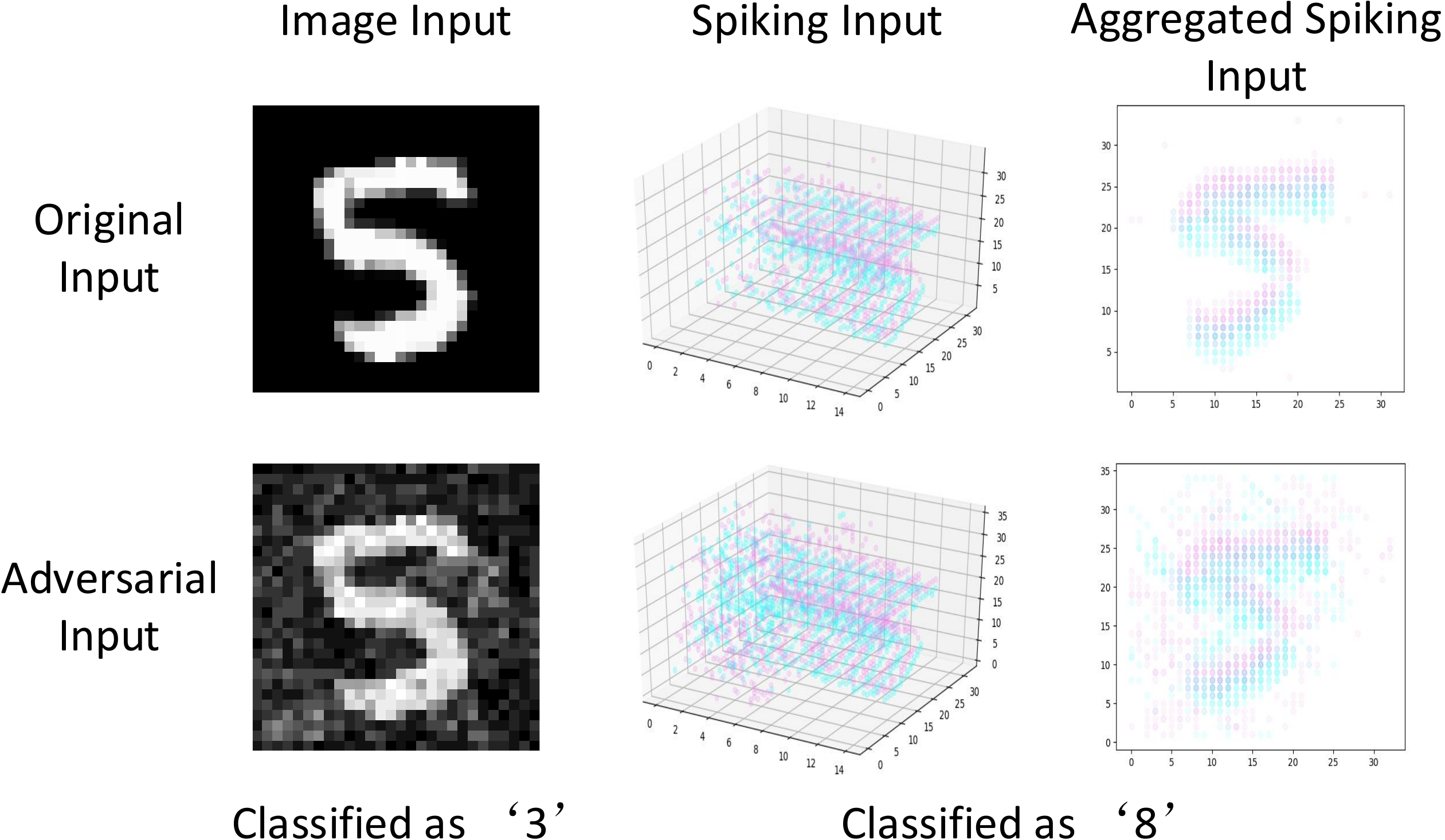}
\caption{The data format of original inputs and adversarial examples. The red and blue colors denote two spike channels induced by dynamic vision sensors \cite{Lichtsteiner2008A,orchard2015converting}.}
\label{fig:input_data_format}
\end{figure}

\textbf{Input Data Format.} It is natural for an SNN model to handle spike signals. Therefore, considering the datasets containing spike events, such as N-MNIST \cite{orchard2015converting} and CIFAR10-DVS \cite{li2017cifar10}, is the first choice. In this case, the input is originally in a spatio-temporal pattern with a binary value for each element (0-nothing; 1-spike). The attacker can flip the state of selected elements, while the binary format must be maintained. Due to the lack of spiking data sources in reality, the image datasets are also widely used in the SNN field by converting them into the spiking version. There are different ways to perform the data conversion, such as rate coding \cite{deng2020rethinking, wu2019direct,sengupta2019going} and latency coding \cite{comsa2020temporal, mostafa2017supervised, kheradpisheh2019s4nn}. In this work, we adopt the former scheme based on Bernoulli sampling that converts the pixel intensity to a spike train (recalling the ``Sample'' in Figure \ref{fig:attack_ann_snn}), where the spike rate is proportional to the intensity value. In this case, the attacker can modify the intensity value of selected pixels by adding the continuous perturbation. Figure \ref{fig:input_data_format} illustrates the adversarial examples in these two cases.

\begin{figure*}[!htbp]
\centering
\includegraphics[width=0.95\textwidth]{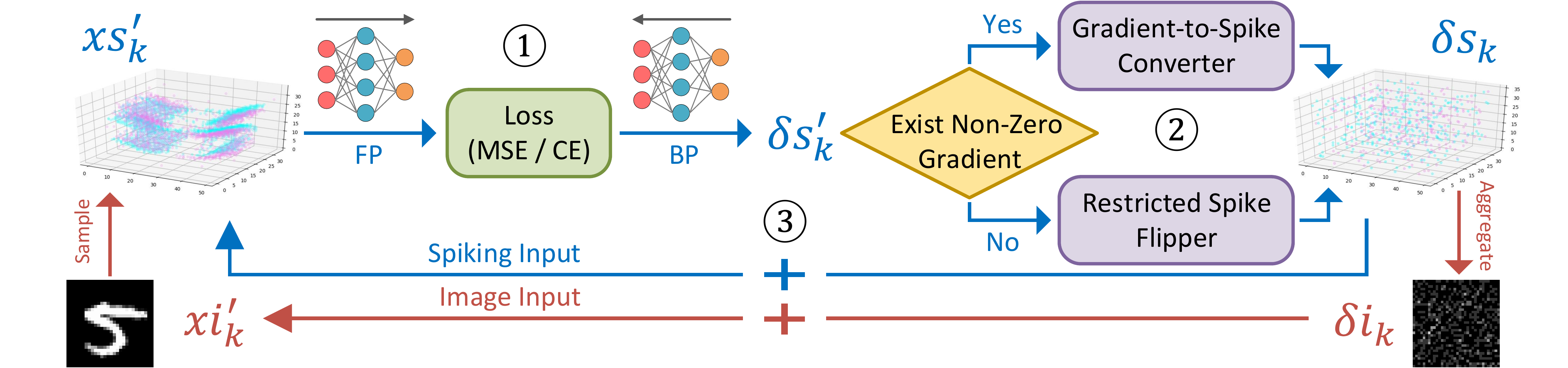}
\caption{Overview of the adversarial attack flow for SNNs with spiking inputs or image inputs. The attack flow consists of: {\textcircled{\normalsize 1}} calculating continuous spatio-temporal input gradients via BPTT; {\textcircled{\normalsize 2}} generating spike-compatible input gradients; {\textcircled{\normalsize 3}} updating adversarial examples. For image-based inputs, an additional aggregation of the input gradients along the temporal dimension is needed.}
\label{fig:attack_flow}
\end{figure*}

\subsection{Attack Flow Overview}

The overview of the proposed adversarial attack against SNNs is illustrated in Figure \ref{fig:attack_flow}. The basic flow adopts the BIM method given in Equation (\ref{equ:bim}), which is the result after considering the spiking inputs of SNNs. The perturbation for spikes can only flip the binary states (0 or 1) of selected input elements rather than add continuous values. Therefore, to generate spiking adversarial examples that are able to cross the decision boundary, the search of candidate elements is more important than the perturbation magnitude. FGSM cannot do this since it only explores the perturbation magnitude, while BIM realizes this by searching new candidate elements in different iterations.

As aforementioned, there are three stages: forward pass (FP), backward pass (BP), and input update, which is executed iteratively until the attack succeeds. The gradients here are in a spatio-temporal pattern, which matches the spatio-temporal dynamics of SNNs and enables a higher attack effectiveness. Besides, the incompatibility between continuous gradients and binary inputs is addressed by the proposed gradient-to-spike (G2S) converter; the gradient vanishing problem is solved by the proposed restricted spike flipper (RSF). Next, we describe the specific flow for spiking inputs and image inputs individually for a clear understanding.

\noindent\textbf{Spiking Inputs.} The blue arrows in Figure \ref{fig:attack_flow} illustrate this case. The generation of spiking adversarial examples relies on three steps as follows. In step {\textcircled{\normalsize 1}}, the continuous gradients are calculated in the FP and BP stages by 
\begin{equation}
    \label{equ:gradient_spike}
    \begin{cases}
    \delta s'_k = \triangledown_{xs'_k} L(\theta, xs_k, y_{original}),~\text{if untargeted}\\
    \delta s'_k = -\triangledown_{xs'_k} L(\theta, xs_k, y_{target}),~\text{if targeted}
    \end{cases}
\end{equation}
where $\delta s'_k$ represents the input gradient map at the $k$-th iteration. Since all elements in $\delta s'_k$ are continuous values, they cannot be directly accumulated onto the spiking inputs (i.e. $xs_k$) to avoid breaking the data format of binary spikes. Therefore, in the step {\textcircled{\normalsize 2}}, we propose G2S converter to convert the continuous gradient map to a ternary one compatible with the spike input, which can simultaneously maintain the input data format and control the perturbation magnitude. When the input gradient vanishes (i.e. all elements in $\delta s'_k$ are zero), we propose RSF to construct a ternary gradient map that can randomly flip the input spikes with a controllable turnover rate. At last, step {\textcircled{\normalsize 3}} accumulates the ternary gradients onto the spiking input.

\noindent\textbf{Image Inputs.} Sometimes, the benchmarking models convert image datasets to spike inputs via Bernoulli sampling. In this case, one more step is additionally needed to generate image-style adversarial examples, which is shown by the red arrows in Figure \ref{fig:attack_flow}. After the above step {\textcircled{\normalsize 2}}, the ternary gradient map should be aggregated in the temporal dimension, i.e. averaging all elements belonging to the same spatial location but in different timesteps, according to $\delta i_k = \frac{1}{T} \sum_{t=1}^T \delta s_k^t$. After this temporal aggregation, the image-compatible input perturbation can be acquired. In each update iteration, the intensity value of $xi_k$ will be clipped within $[0,~1]$. Note that, although our method aggregates the gradients from all timesteps in this scenario, the gradient at every timestep is directly found in the target SNN model, which can be more accurate than those found by prior approaches.

\subsection{Acquisition of Spatio-Temporal Gradients}

We introduce the state-of-the-art supervised learning algorithms for SNNs \cite{wu2018spatio,wu2019direct,bellec2018long}, which are inspired by the backpropagation through time (BPTT) to acquire the gradients in both spatial and temporal dimensions. Here we take the one in \cite{wu2019direct} as an illustrative example. In order to simulate in current programming frameworks (e.g. Pytorch), the original LIF neuron model in Equation (\ref{equ:lif}) should be first converted to its equivalent iterative version. Specifically, we have
\begin{equation}
    \label{equ:lif_iterative}
    \begin{cases}
    u^{t+1, n+1}_i&= e^{-\frac{dt}{\tau}}u_i^{t, n+1}(1-o_i^{t,n+1}) + \sum_jw_{ij}^no_j^{t+1, n} \\
    o^{t+1, n+1}_i&= fire(u_i^{t+1, n+1}-u_{th})
    \end{cases}
\end{equation}
where $t$ and $n$ represent indices of the simulation time step and the layer, respectively, $dt$ is the time step length, and $e^{-\frac{dt}{\tau}}$ reflects the leakage effect of the membrane potential. $fire(\cdot)$ is a step function, which satisfies $fire(x)=1$ when $x\geq 0$, otherwise $fire(x)=0$. This iterative LIF model incorporates all behaviors of a spiking neuron, including integration, fire, and reset. Note that a spike can be simply modeled as a binary event (1 or 0) in the above discrete time domain, which differs from that in the continuous time domain in Equation (\ref{equ:lif}).

In the output layer, we adopt the commonly-used spike rate coding scheme for the recognition, i.e., the neuron firing the most spikes becomes the winner that indicates the class prediction. The spatio-temporal spike pattern of the output layer is converted into a spike rate vector, described as 
\begin{equation}
\label{equ:spike_rate}
y_i=\frac{1}{T} \sum_{t=1}^T  o_i^{t,N}
\end{equation}
where $N$ is the output layer index and $T$ is the length of the simulation time window. This spike rate vector can be regarded as the normal output vector in ANNs. With this output conversion, the typical loss functions $L$ for ANNs, such as mean square error (MSE) and cross-entropy (CE), can also be applied in the loss function for SNNs.

\begin{figure}[!htbp]
\centering
\includegraphics[width=0.25\textheight]{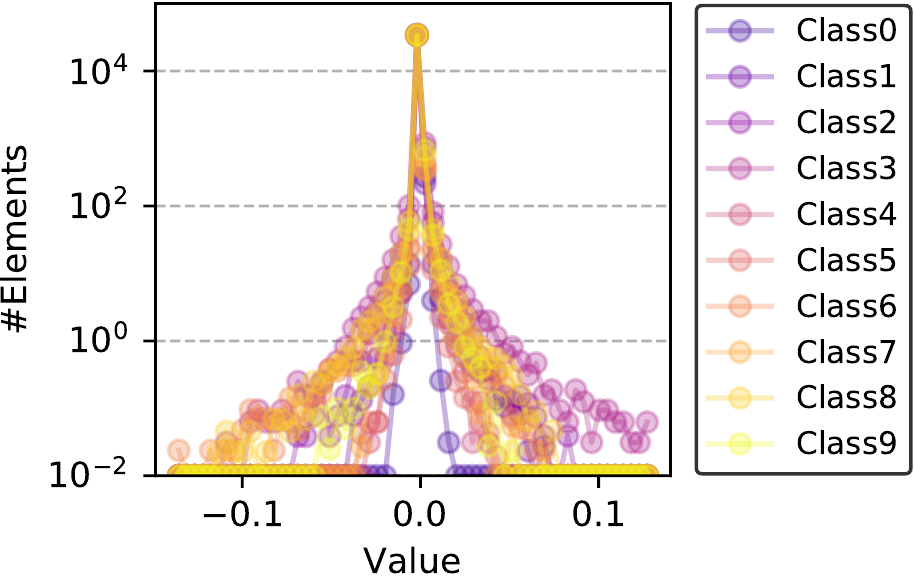}
\caption{The distribution of input gradients overall 500 samples from N-MNIST. The model is trained with MSE loss. Most of gradients are zero which might lead to the gradient vanishing problem.}
\label{fig:distribution}
\end{figure}

Based on the iterative LIF neuron model and a given loss function, the gradient propagation can be governed by 
\begin{equation}
\begin{cases}
\label{equ:stbp}
\frac{\partial L}{\partial o_i^{t,n}}
=
\sum_{j}\frac{\partial L}{\partial u_j^{t,n+1}}\frac{\partial u_j^{t,n+1}}{\partial o_i^{t,n}} + \frac{\partial L}{\partial u_i^{t+1,n}}\frac{\partial u_i^{t+1,n}}{\partial o_i^{t,n}}
\\
\frac{\partial L}{\partial u_i^{t,n}}
=
\frac{\partial L}{\partial o_i^{t,n}} \frac{\partial o_i^{t,n}}{\partial u_i^{t,n}}+\frac{\partial L}{\partial u_i^{t+1,n}}\frac{\partial u_i^{t+1,n}}{\partial u_i^{t,n}}
\end{cases}.
\end{equation}
However, the firing function is non-differentiable, i.e. $\frac{\partial o}{\partial u}$ does not exist. As mentioned earlier, a Dirac-like function is introduced to approximate its derivative \cite{wu2018spatio}. Specifically, $\frac{\partial o}{\partial u}$ can be estimated by
\begin{equation}
    \label{equ:o_u}
    \frac{\partial o}{\partial u} \approx
    \begin{cases}
    \frac{1}{a},~~|u-u_{th}|\leq \frac{a}{2} \\
    0,~~otherwise
    \end{cases}
\end{equation}
where $a$ is a hyper-parameter to control the gradient width when passing the firing function during backpropagation. This approximation indicates that only the neurons whose membrane potential is close to the firing threshold have the chance to let gradients pass through, as shown in Figure \ref{fig:distribution}. It can be seen that abundant zero gradients are produced, which might lead to the gradient vanishing problem (all input gradients become zero). 

During the adversarial attack, we consider the attack is successful when the adversarial example can fool the SNN model still under the spike rate coding scheme, i.e., letting any other neuron fire the most spikes (untargeted attack) or letting the targeted neuron fire the most spikes (targeted attack).

\subsection{Gradient-to-Spike (G2S) Converter}\label{sec:app:G2S}

\begin{figure}[!htbp]
\centering
\includegraphics[width=0.48\textwidth]{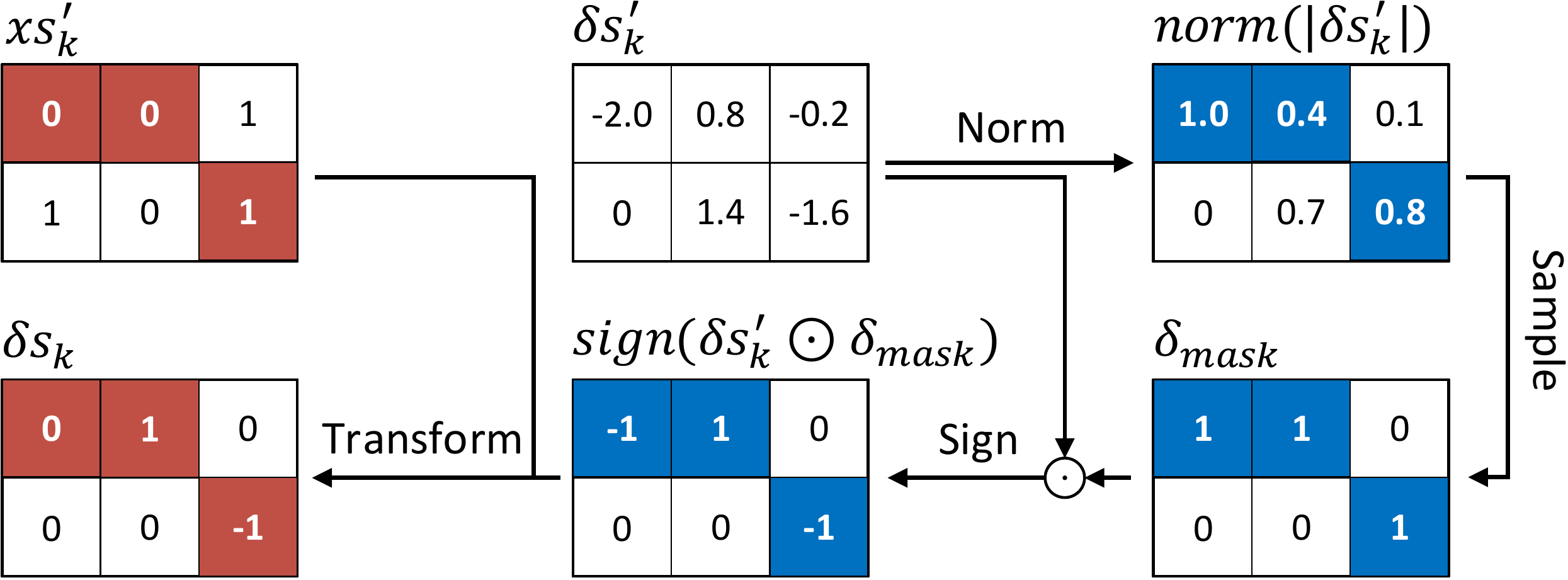}
\caption{Illustration of gradient-to-spike (G2S) converter with probabilistic sampling reducing the number of modified points and thus lowering the perturbation magnitude, sign extraction ternarizing the continuous gradients for spike compatibility, and overflow-aware transformation clipping the data range in adversarial examples.}
\label{fig:g2s}
\end{figure}

There are two goals in the design of G2S converter in each iteration: (1) the final gradients should be compatible with the spiking inputs, i.e. remaining the spike format unchanged after the gradient accumulation; (2) the perturbation magnitude should be imperceptible, i.e. limiting the number of non-zero gradients. To this end, we design three steps: probabilistic sampling, sign extraction, and overflow-aware transformation, which are illustrated in Figure \ref{fig:g2s}.

\begin{figure}[!htbp]
\centering
\includegraphics[width=0.38\textwidth]{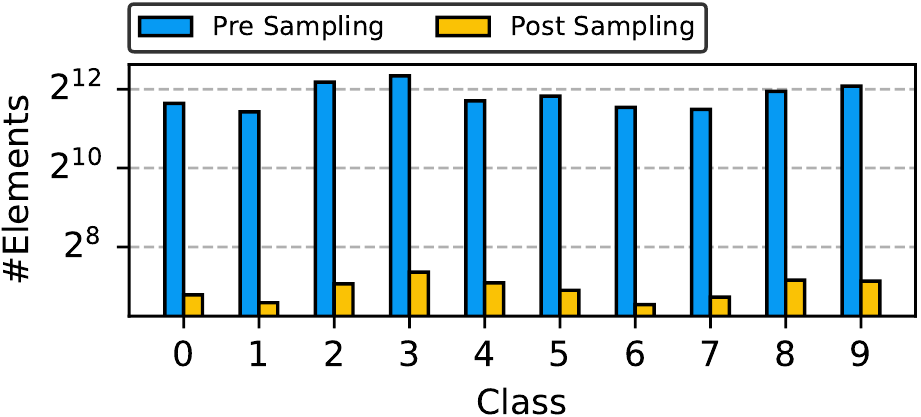}
\caption{The number of elements with non-zero input gradients before and after the probabilistic sampling. We report the average data across the inputs in each class. After the probabilistic sampling step, the number of selected non-zero input elements for modification is reduced a lot.}
\label{fig:nonzero_gradient}
\end{figure}

\noindent\textbf{Probabilistic Sampling.} The absolute value of the input gradient map obtained by Equation (\ref{equ:gradient_spike}), i.e. $|\delta s'_k|$, is first normalized to lie in the range of $[0,~1]$. Then, the normalized gradient map, i.e. $norm(|\delta s'_k|)$, is sampled to produce a binary mask with the same shape, in which the 1s indicate the locations where gradients can pass through. The probabilistic sampling for each gradient element obeys 
\begin{equation} \label{equ:mask}
\begin{cases}
    P(\delta_{mask}=1) = norm(|\delta s'_k|)\\
    P(\delta_{mask}=0)= 1 - norm(|\delta s'_k|)
\end{cases}.
\end{equation}
In other words, a larger gradient has a larger possibility to let the gradient pass through. By multiplying the resulting mask with the original gradient map, the number of non-zero elements can be reduced significantly. To evidence this conclusion, we run the attack against the SNN model with a network structure to be provided in Table \ref{tab:net_structure} over 500 spiking inputs from N-MNIST, and the results are presented in Figure \ref{fig:nonzero_gradient}. Given MSE loss and untarget attack scenario, the number of non-zero elements in $\delta s'_k$ could reach $2^{10}$. After using the probabilistic sampling, the number of non-zero elements in $\delta s'_k \odot \delta_{mask}$ can be greatly decreased, masking out $>96\%$ percentage.

\noindent\textbf{Sign Extraction.} Now, we explain how to generate a ternary gradient map where each element is in $\{-1,~0,~1\}$, which can maintain the spike format after accumulating onto the spiking inputs with binary values of $\{0,~1\}$. This step is simply based on a sign extraction:
\begin{equation} \label{equ:delta_mask}
    \delta s_k'' = sign(\delta s'_k \odot \delta_{mask})
\end{equation}
where we define $sign(x)=1$ if $x>0$, $sign(x)=0$ if $x=0$, and $sign(x)=-1$ otherwise.  

\noindent\textbf{Overflow-aware Transformation.} Although the above $\delta s_k''$ is able to be ternary, it cannot ensure that the final adversarial example generated by input gradient accumulation is still limited in $\{0,~1\}$. For example, an original ``0'' element in $xs_k$ with a ``$-$1'' gradient or an original ``1'' element with a ``1'' gradient will yield a ``$-$1'' or ``2'' input that is out of  $\{0,~1\}$. This overflow breaks the data format of binary spikes. To address this issue, we propose an overflow-aware gradient transformation to constrain the range of the final adversarial example, which is illustrated in Table \ref{tab:overflow_transform}.

\begin{table}[!htbp]
\caption{Overflow-aware gradient transformation.}
\label{tab:overflow_transform}
\vspace{3pt}
\centering
\renewcommand\arraystretch{1.3}
\resizebox{0.42\textwidth}{!}{
\begin{tabular}{c | c c | c c }   
\hline \hline
& \multicolumn{2}{c|}{Before Transformation} & \multicolumn{2}{c}{After Transformation}\\
   $xs'_k$ & $\delta s_k''$ & $xs_k+\delta s_k''$ & $\delta s_k$ & $xs'_k+\delta s_k$ \\
    \hline
    0/1 & 0 & 0/1 & 0 & 0/1\\
      0 & 1 & 1   & 1 & 1\\
      1 & 1 & 2   & 0 & 1\\
      0 & -1 & -1 & 0 & 0\\
      1 & -1 & 0 & -1 & 0\\
\hline \hline
\end{tabular}}
\end{table}

After introducing the above three steps, now the function of G2S converter can be briefly summarized as below:
\begin{equation} \label{equ:g2s}
    \delta s_k = transform[sign(\delta s'_k \odot \delta_{mask}),~xs'_k]
\end{equation}
where $transform(\cdot)$ denotes the overflow-aware transformation. The G2S converter is able to achieve the two goals mentioned earlier by simultaneously keeping the spike compatibility and controlling the perturbation magnitude.

\subsection{Restricted Spike Flipper (RSF)}

\begin{figure}[!htbp]
\centering
\includegraphics[width=0.32\textwidth]{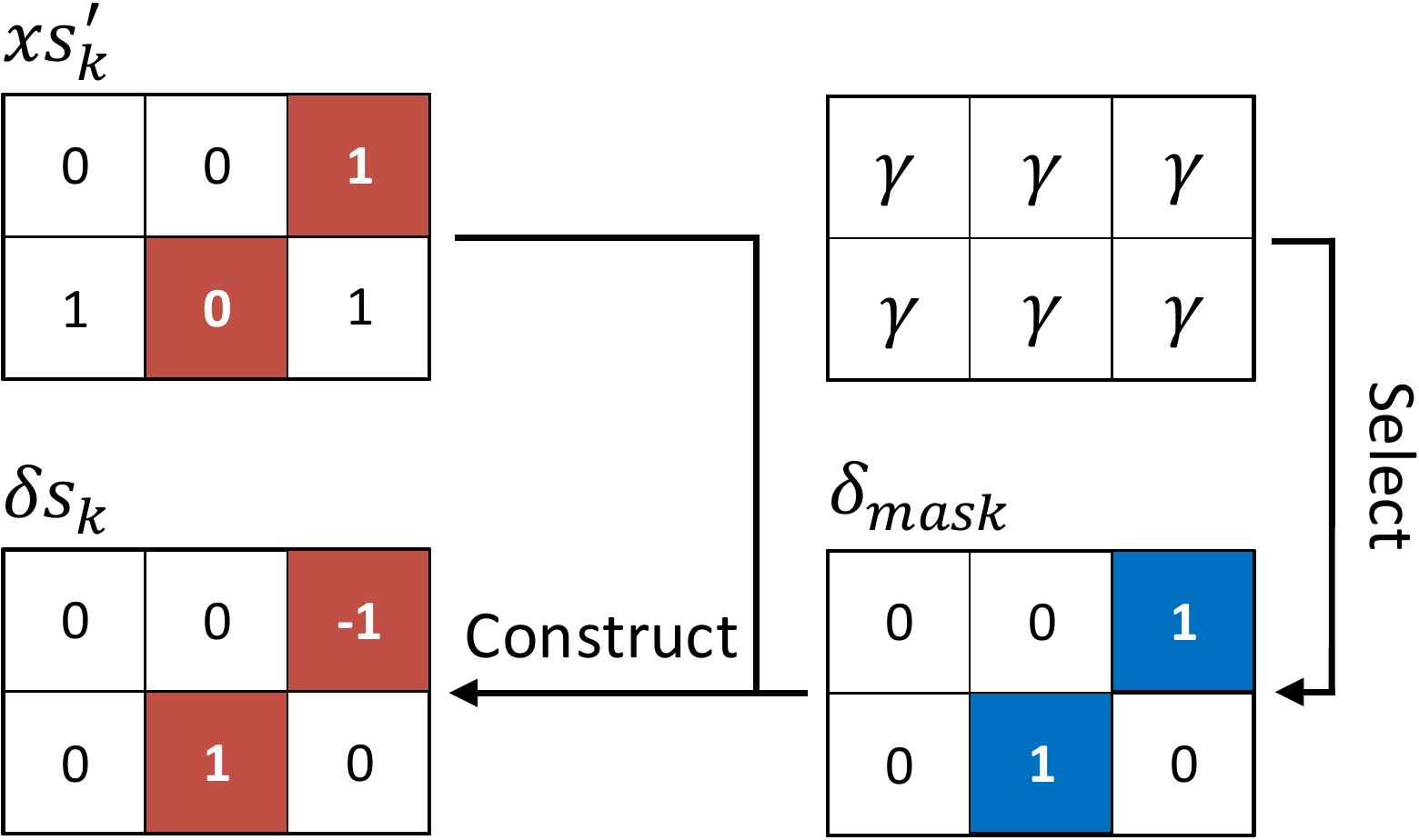}
\caption{Illustration of restricted spike flipper (RSF) with element selection reducing the number of modified points through probabilistic sampling over a pre-defined probability $\gamma$ and gradient construction creating spike-compatible gradients through spike flipping.}
\label{fig:gt}
\end{figure}

Table \ref{tab:gradient_vanishing} identifies the gradient vanishing issue in SNNs trained by BPTT, which is quite severe. One way to mitigate the gradient vanishing problem is increasing the parameter $a$ in Equation \ref{equ:o_u}, which allows more neurons receiving gradients during the backward pass. However, a too large $a$ might lead to inaccuracy when approximating the gradient of the firing function, therefore the gradient vanishing problem cannot be fully resolved by increasing $a$. To this end, we design RSF to solve the gradient vanishing problem by constructing gradients artificially. The constraints of the constructed gradient map are the same with those of G2S converter, i.e. spike format compatibility and perturbation magnitude controllability. To this end, we design two steps: element selection and gradient construction, which is illustrated in Figure \ref{fig:gt}.

\begin{table}[!htbp]
\caption{Number of inputs with all-zero gradients at the first attack iteration. We test the untargeted attack with over 500 inputs for each dataset. }
\vspace{3pt}
\label{tab:gradient_vanishing}
\centering
\resizebox{0.48\textwidth}{!}{
\begin{tabular}{c | c | c | c | c }   
\hline \hline
Dataset & N-MNIST & CIFAR10-DVS & MNIST & CIFAR10 \\
\hline
\#grad.-vanish. inputs (MSE) & 130 & 41 & 436 & 103  \\
\#grad.-vanish. inputs (CE)~ & 256 & 32 & 471 & 105\\ 
\hline \hline
\end{tabular}}
\end{table}

\noindent\textbf{Element Selection.} This step is to select the elements to let gradients pass through. In G2S converter, the probabilistic sampling is used to produce a binary mask to indicate the element selection, whereas, all the gradients in $\delta s'_k$ are zeros here. To continue the use of the above probabilistic sampling method, we provide a gradient initialization that sets all elements to $\gamma$ as the example provided in Figure \ref{fig:gt}. $\gamma$ is a factor within the range of [0, 1], which controls the number of non-zero gradients after RSF. Now the probabilistic sampling in Equation (\ref{equ:mask}) is still applicable to generate the mask $\delta_{mask}$.


\begin{table}[!htbp]
\caption{Gradient construction to flip spiking inputs.}
\label{tab:gradient_gen}
\vspace{2pt}
\centering
\renewcommand\arraystretch{1.3}
\resizebox{0.27\textwidth}{!}{
\begin{tabular}{c c | c c  }   
\hline \hline
& & \multicolumn{2}{c}{After Construction}\\
   $xs'_k$ & $\delta_{mask}$ & $\delta s_k$ & $xs'_k+\delta s_k$ \\
    \hline
    0/1 & 0 & 0 & 0/1\\
      0 & 1 & 1 & 1\\
      1 & 1 & -1  & 0\\
\hline \hline
\end{tabular}}
\end{table}

\noindent\textbf{Gradient Construction.} To maintain the spike format of adversarial examples, we just flip the state of spiking inputs in the selected region. Here the flipping means switching the element state to ``0'' if it is ``1'' currently, or vice versa. Table \ref{tab:gradient_gen} illustrates the construction of ternary gradients that are able to flip the spiking inputs. 

With the above two steps, the spiking inputs can be flipped randomly with a good control of the overall turnover rate. The overall function of RSF can be expressed as
\begin{equation} \label{equ:gt}
    \delta s_k = construct(\delta_{mask},~xs'_k).
\end{equation}
To summarize, RSF continues the update of adversarial examples interrupted by the gradient vanishing.

\subsection{Overall Attack Algorithm}

Based on the explanations of G2S converter and RSF, Algorithm \ref{alg:overall_attack} provides the overall attack algorithm corresponding to the attack flow illustrated in Figure \ref{fig:attack_flow}. For different input data formats, we give different ways to generate adversarial examples. There are several hyper-parameters in our algorithm, such as the maximum attack iteration number ($Iter$), the norm format ($p$) to quantify the perturbation magnitude, the perturbation magnitude upper bound ($\epsilon$), the gradient scaling rate ($\eta$), and the sampling factor ($\gamma$) in RSF. Notice that we use the average perturbation per point as the metric to evaluate the perturbation magnitude in each adversarial example with $N$ pixel points, i.e., $\frac{1}{N} \lVert x'_{k+1} - x'_0 \rVert_p$.

\begin{algorithm}
\label{alg:overall_attack}
\caption{The overall SNN attack algorithm.}
\textbf{Input}: $x$, $Iter$, $p$, $\epsilon$, $\eta$, $\gamma$;\\
\textbf{if} $image ~ input$ \textbf{then} $xi_0 = x$; \textbf{end}\\
\textbf{else} $xs'_0 = x$; \textbf{end}\\

\BlankLine
\For{$k = 1 ~ to ~ Iter$}{
    \If{image input}{
    $xs'_k \leftarrow$ Bernoulli sampling on $xi'_k$;\\
    }
    \BlankLine
    Get $\delta s'_k$ through Equation (\ref{equ:gradient_spike});\\
    \BlankLine
    \If{gradient vanishing occurs in $\delta s'_k$}{ 
        // RSF \\
        $\delta_{mask} \leftarrow$ Probabilistic sampling on $\gamma$; \\
        $\delta s_k = construct(\delta_{mask},~xs'_k)$;\\
    }
    \Else{ 
        // G2S converter \\
        $\delta_{mask} \leftarrow$ Probabilistic sampling on $norm(|\delta s'_k|)$; \\
        $\delta s_k = transform[sign(\delta s'_k \odot \delta_{mask}),~xs'_k]$;\\
    }
    \BlankLine
    \If{image input}{
    $\delta i_k \leftarrow \frac{1}{T} \sum_{t=1}^T \delta s_k^t$; // Temporal aggregation \\

    \BlankLine
    $xi'_{k+1}=xi'_k + \delta i_k$; \\
    \BlankLine
    \If{$\frac{1}{N} \lVert xi'_{k+1} - xi'_0 \rVert_p \geqslant \epsilon$}{
       \BlankLine
       \textbf{break}; // Attack failed
    }
    \BlankLine
    \If{attack succeeds}{
        \textbf{return} $xi'_{k+1}$; // Attack successful
    }
    }
    \Else{
    $xs'_{k+1}=xs'_k + \delta s_k$; \\
    \BlankLine
    \If{$\frac{1}{N} \lVert xs'_{k+1} - xs'_0 \rVert_p \geqslant \epsilon$}{
       \BlankLine
       \textbf{break}; // Attack failed
    }
    \BlankLine
    \If{attack succeeds}{
        \textbf{return} $xs'_{k+1}$; // Attack successful
    }
    }
}
\BlankLine
\end{algorithm}

\section{Loss Function and Firing Threshold}\label{sec:loss_thre}
\label{sec: knobs}
In this work, we consider two design knobs that affect the SNN attack effectiveness: the loss function during training and the firing threshold of the penultimate layer during attack.

\subsection{MSE and CE Loss Functions}

\begin{figure}[!htbp]
\centering
\includegraphics[width=0.48\textwidth]{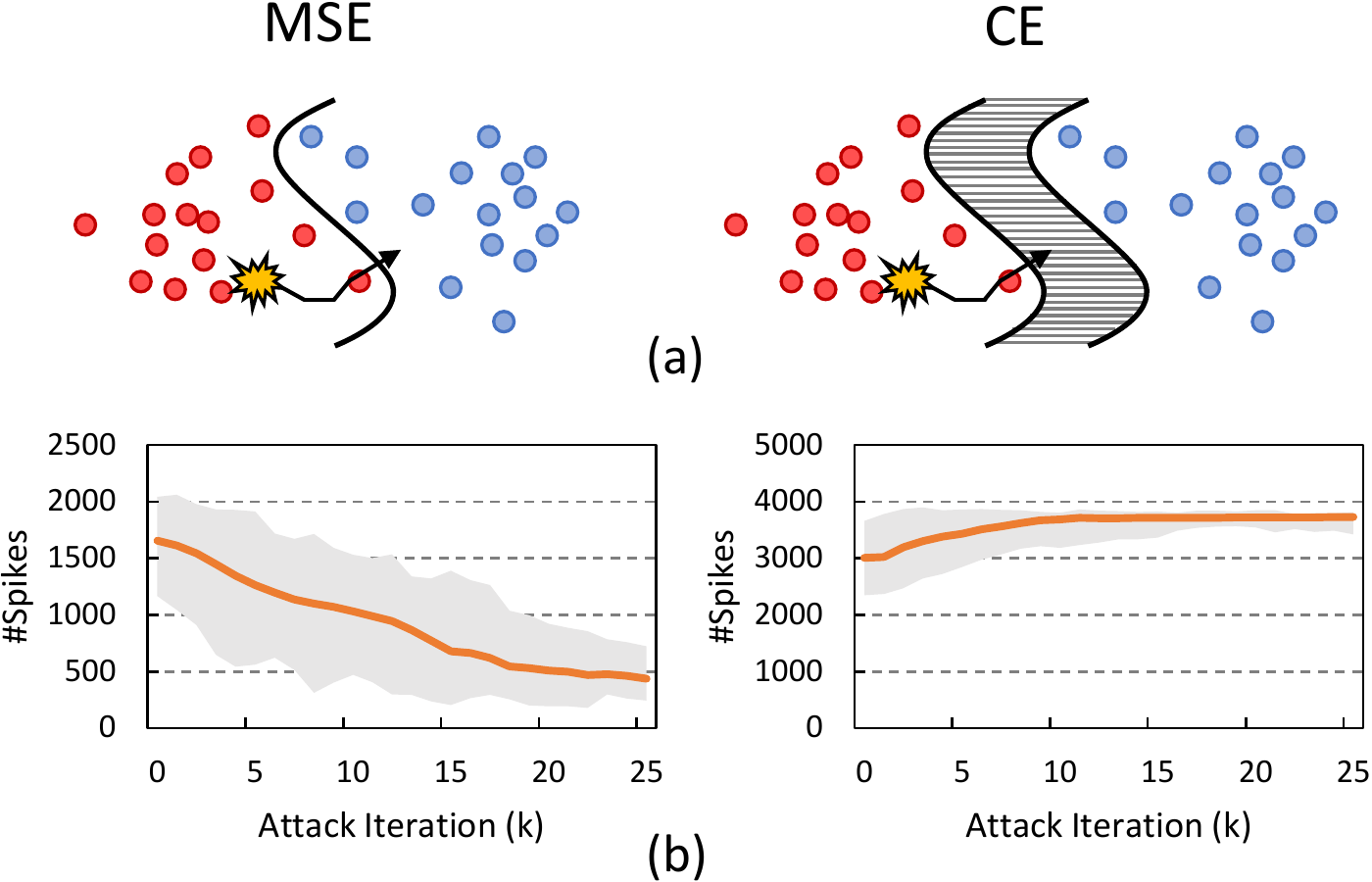}
\caption{Loss function analysis: (a) decision boundary comparison; (b) the number of output spikes in the penultimate layer at different attack iterations. We report the average data across different inputs. The shaded area in (a) represents a ``trap'' area that cannot be resolved by RSF; the larger number of spikes in the penultimate layer under CE probably introduces the ``trap'' effect.}
\label{fig:loss}
\end{figure}

We compare two widely used loss functions, mean square error (MSE) loss and cross entropy (CE) loss. The former directly receives the fire rate of the output layer, while the latter needs an extra softmax layer following the output fire rate. We observe that the gradient vanishing occurs more often when the model is trained by CE loss. It seems that there is a ``trap'' region in this case, which means the output neurons cannot change the response any more no matter how RSF modifies the input. We use Figure \ref{fig:loss}(a) to illustrate our finding. When we use CE loss during training, the gradient is usually vanished between the decision boundaries (i.e. the shaded area) and cannot be recovered by RSF; while this phenomenon seldom happens if MSE loss is used.

For a deeper understanding, we examine the output pattern of the penultimate layer (during untargeted attack) since it directly interacts with the output layer, as depicted in Figure \ref{fig:loss}(b). Here the network structure will be provided in Table \ref{tab:net_structure} and the 500 test inputs are randomly selected from the N-MNIST dataset. When the training loss is MSE, the number of output spikes in the penultimate layer gradually decreases as the attack process evolves. On the contrary, the spike number first increases and then stays unchanged for the CE trained model. Based on this observation, one possible hypothesis is that more output spikes in the penultimate layer might increase the distance between decision boundaries, thus introducing the mentioned ``trap'' region with gradient vanishing.

\begin{figure}[!htbp]
\centering
\includegraphics[width=0.48\textwidth]{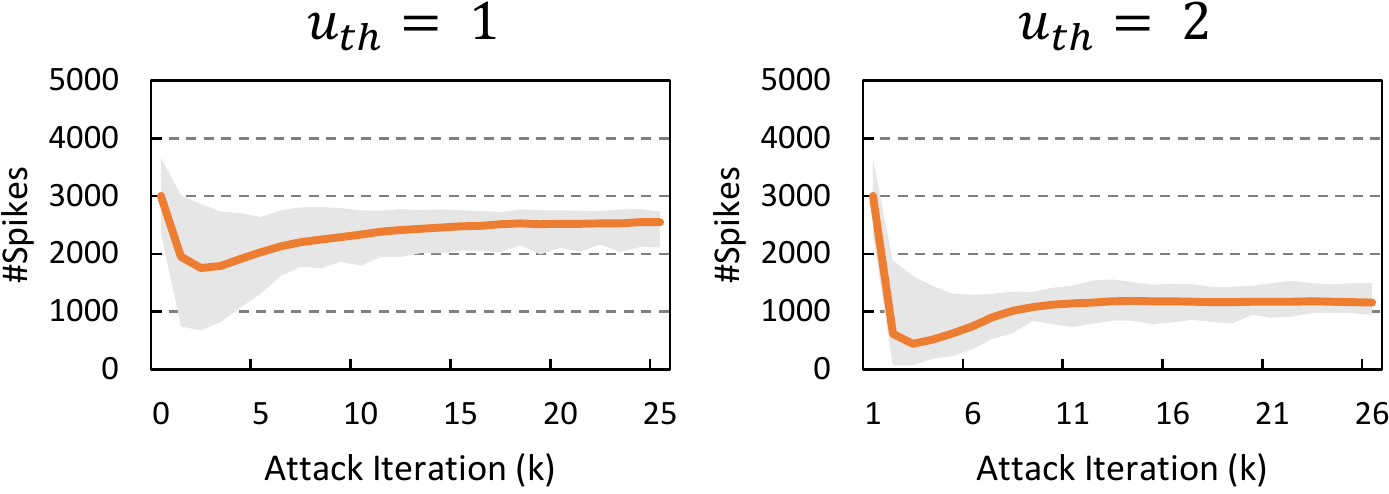}
\caption{The number of output spikes in the penultimate layer with different firing threshold in that layer. We report the average data across different inputs. The increase of firing threshold in the penultimate layer is able to reduce the number of spikes.}
\label{fig:thresh}
\end{figure}

\subsection{Firing Threshold of the Penultimate Layer}

As introduced in the above subsection, the models trained by CE loss are prone to output more spikes in the penultimate layer, leading to the ``trap'' region that makes the attack difficult. To address this issue, we increase the firing threshold of the penultimate layer during attack to reduce the number of spikes there. Notice that we only modify the firing threshold in the FP stage during the generation of adversarial examples (see Figure \ref{fig:attack_flow}). The original model itself is kept unchanged when facing attack with the generated adversarial examples, thus the threshold tuning does not affect the network accuracy. With the threshold tuning, we present the number of spikes again in Figure \ref{fig:thresh}, where the CE loss is used and other settings are the same with those in Figure \ref{fig:loss}(b). Compared to the original threshold setting ($u_{th}=0.3$) in the previous experiments, the number of output spikes in the penultimate layer can be decreased significantly on average. Latter experiments in Section \ref{sec:exp:loss_threshold} will evidence that this tuning of firing threshold is able to improve the adversarial attack effectiveness.

%% file: text/Results.tex
\section{Experiment Results}\label{sec:Results}

\subsection{Experiment Setup}

We design our experiments on both spiking and image datasets. The spiking datasets include N-MNIST \cite{orchard2015converting} and CIFAR10-DVS \cite{li2017cifar10} which are captured by dynamic vision sensors \cite{Lichtsteiner2008A}; while the image datasets include MNIST \cite{lecun1998gradient} and CIFAR10 \cite{krizhevsky2009learning}. For these two kinds of dataset, we use different network structure, as listed in Table \ref{tab:net_structure}. For each dataset, the detailed hyper-parameter setting during training and the trained accuracy are shown in Table \ref{tab:para_acc}. The default loss function is MSE unless otherwise specified. Note that since we focus on the attack methodology in this work, we do not use the optimization techniques such as input encoding layer, neuron normalization, and voting-based classification \cite{wu2019direct} to improve the training accuracy.

\begin{table}[!htbp]
\caption{Network structure on  different datasets. ``C'', ``AP'', and ``FC'' denote convolutional layer, average pooling layer, and fully-connected layer, respectively.}
\vspace{3pt}
\label{tab:net_structure}
\centering
\renewcommand\arraystretch{1.3}
\resizebox{0.48\textwidth}{!}{
\begin{tabular}{c | c }   
    \hline
    \hline
    Dataset & Network Structure \\
    \hline
    Spike & Input-128C3-128C3-AP2-384C3-384C3-AP2-1024FC-512FC-10FC \\
    \hline
    Image & Input-128C3-256C3-AP2-512C3-AP2-1024C3-512C3-1024FC-512FC-10FC \\
    \hline
    Gesture-DVS & Input-64C3-128C3-AP2-128C3-AP2-256FC-11FC \\
    \hline
    \hline
\end{tabular}}
\end{table}

\begin{table}[!htbp]
\caption{Hyper-parameter settings and model accuracy during training.}
\vspace{3pt}
\label{tab:para_acc}
\centering
\renewcommand\arraystretch{1.3}
\resizebox{0.48\textwidth}{!}{
\begin{tabular}{c | c | c | c | c | c}   
    \hline
    \hline
     Datasets & Gesture-DVS & N-MNIST & CIFAR10-DVS & MNIST & CIFAR10 \\
     \hline
     Input Size & $32\times32\times2$& $34\times34\times2$ & $42\times42\times2$ & $28\times28\times1$ & $32\times32\times3$ \\
     $u_{th}$ & 0.3 & 0.3 & 0.3 & 0.3 & 0.3 \\
     $e^{-\frac{dt}{\tau}}$  & 0.3 & 0.3 & 0.3 & 0.25 & 0.25 \\
     $a$ & 0.5 & 0.5 & 0.5 & 1 & 1 \\
     $T$ & 60 & 15 & 10 & 15 & 15 \\
     Time Bin & 1ms & 5ms & 5ms & - & - \\
     \hline
     Acc (MSE) & 91.32\% & 99.49\% & 64.60\% & 99.27\% & 76.37\%  \\
     Acc (CE) & - &  99.42\% & 64.50\% & 99.52\% & 77.27\% \\
    \hline
    \hline
\end{tabular}}
\end{table}

We set the maximum iteration number of adversarial attack, i.e. $Iter$ in Algorithm \ref{alg:overall_attack}, to 25. We randomly select 50 inputs in each of the 10 classes for untargeted attack and 10 inputs in each class for targeted attack. In targeted attack, we set the target to all classes except the ground-truth one. We use attack success rate and average perturbation perpoint (i.e. $\lVert \delta \rVert_p$) as two metrics to evaluate the attack effectiveness. Specifically, the attack success rate is calculated in the same way as the prior work do~\cite{su2018robustness,goodfellow2014explaining}: for untargeted attack, it is the percentage of the cases that adversarial examples fool the model to output a different label from the ground-truth one; for targeted attack, it is the percentage of the cases that adversarial examples manipulate the model to output the target label. Noted, during the calculation of attack success rate, we only consider the original images that can be correctly classified to eliminate the impact of intrinsic model prediction errors. In the perturbation calculation, we adopt L2 norm, i.e., $p=2$. For image-based datasets, we normalize each input value into $[0, 1]$.

\begin{figure}[!htbp]
\centering
\includegraphics[width=0.47\textwidth]{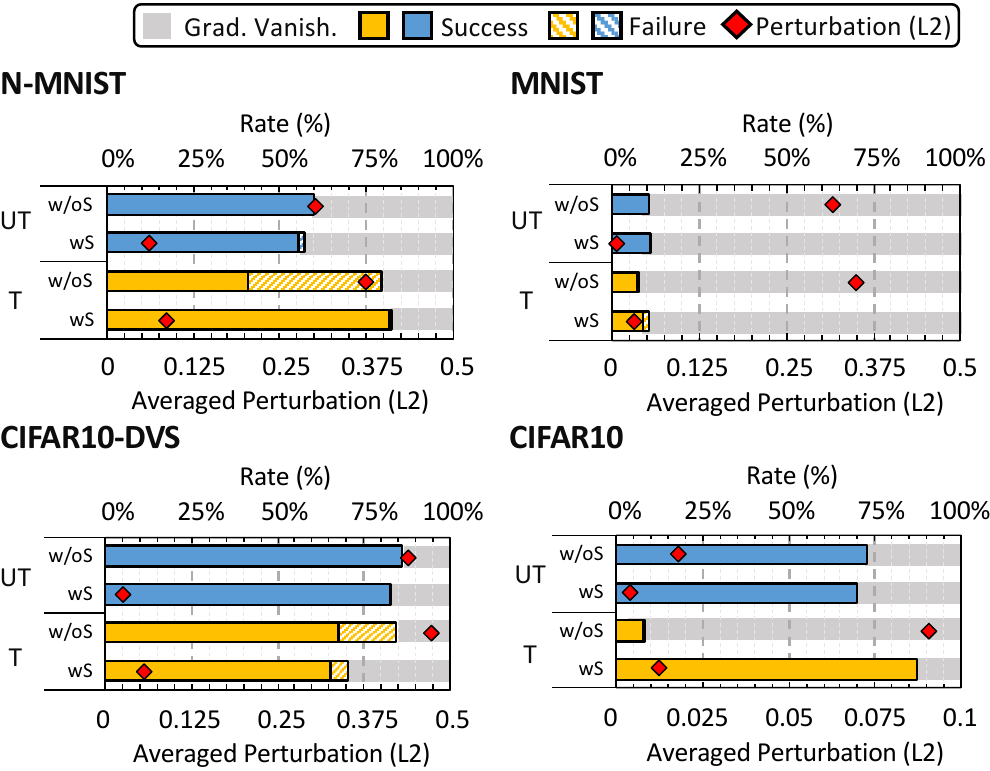}
\caption{Comparison of attack success rate and average perturbation over different datasets with and without probabilistic sampling in G2S converter. ``T'', ``UT'', ``w/oS'', and ``wS'' refer to targeted attack, untargeted attack, G2S without probabilistic sampling, and G2S with probabilistic sampling, respectively. Results indicate that the probabilistic sampling technique can significantly lower the average perturbation.}
\label{res:sampling}
\end{figure}

\subsection{Influence of G2S Converter}\label{sec:res:G2S}

We first validate the effectiveness of G2S converter. Among the three steps in G2S converter (i.e. probabilistic sampling, sign extraction, and overflow-aware transformation) as introduced in Section \ref{sec:app:G2S}, the last two are needed in addressing the spike compatibility while the first one is just used to control the perturbation amplitude. Therefore, we examine how does the probabilistic sampling in G2S converter affects the attack effectiveness. Please note that we do not use RSF to solve the gradient vanishing in this subsection. 

Figure \ref{res:sampling} presents the comparison of attack results (e.g. success/failure rate, gradient vanishing rate, and perturbation amplitude) over four datasets with or without the probabilistic sampling. Both the untargeted attack and the targeted attack are tested. We provide the following several observations. First, the required perturbation amplitude of targeted attack is higher than that of untargeted attack, and the success rate of targeted attack is usually lower than that of untarget attack. These results reflect the difficulty of targeted attack that needs to move the output to an expected class accurately. Second, the probabilistic sampling can significantly reduce the perturbation amplitude in all cases because it removes many small gradients. Third, the success rate (especially of the more difficult targeted attack) can be improved to a great extent on most datasets via the sampling optimization owing to the improved attack convergence with smaller perturbation. With the probabilistic sampling, the attack failure rate could be reduced to almost zero if the gradient is not vanished.

\begin{figure}[!htbp]
\centering
\includegraphics[width=0.47\textwidth]{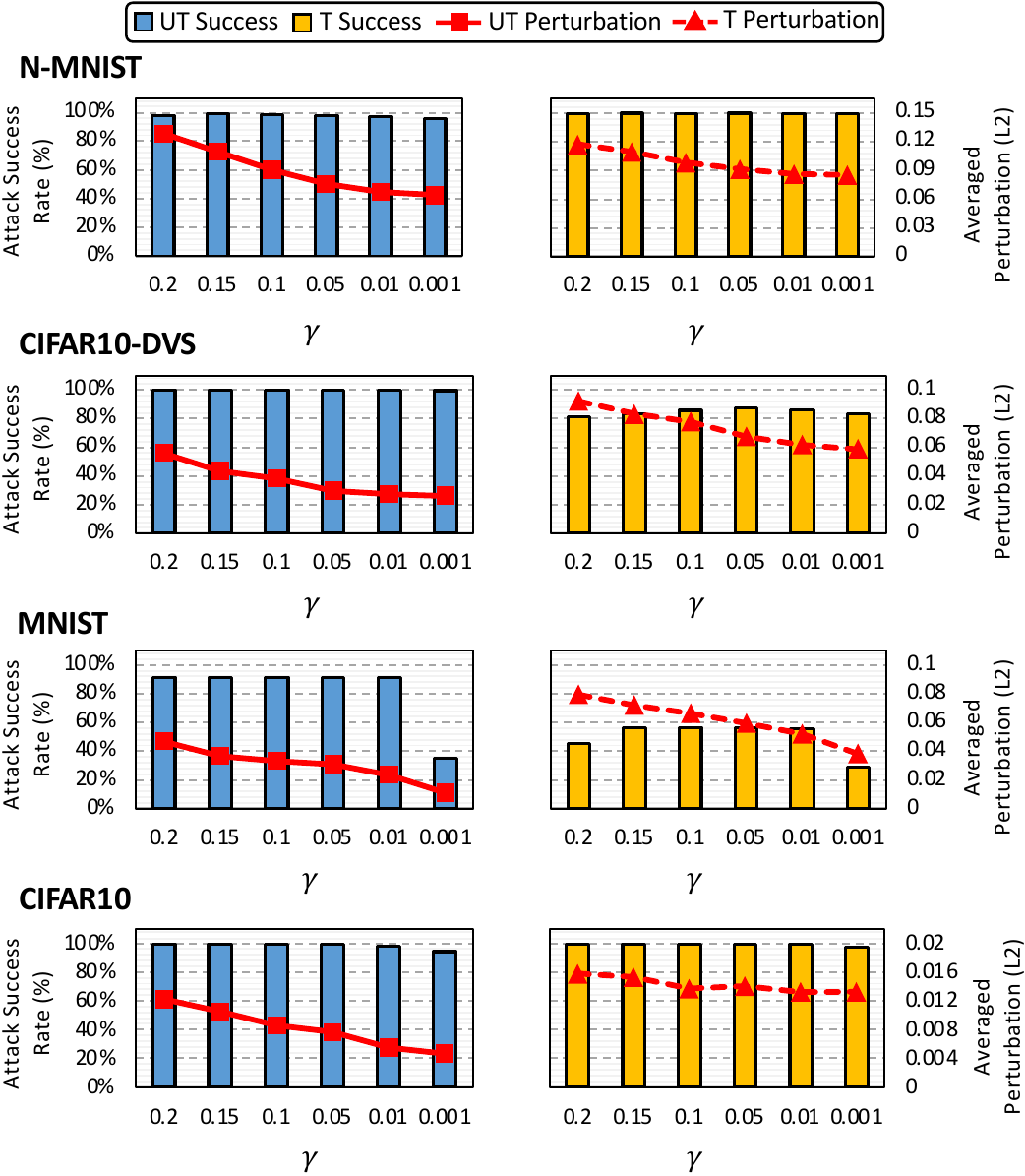}
\caption{Attack success rate and average perturbation with different $\gamma$ settings. ``T'' and ``UT'', refer to targeted attack and untargeted attack, respectively. An appropriate $\gamma$ setting is the best since a too large $\gamma$ will increase the perturbation and might cause a non-convergent attack while a too small $\gamma$ cannot solve the gradient vanishing problem.   
}
\label{fig:gt_result1}
\vspace{-20pt}
\end{figure}

\subsection{Influence of RSF}
Then, we validate the effectiveness of RSF. In RSF, the hyper-parameter $\gamma$ controls the number of selected elements, thus affecting the perturbation amplitude. Keep in mind that a larger $\gamma$ indicates a larger perturbation via flipping the state of more elements in the spiking input. 

We first analyze the impact of $\gamma$ on the attack success rate and perturbation amplitude, as shown in Figure \ref{fig:gt_result1}. A similar conclusion as observed in Section \ref{sec:res:G2S} also holds, that the target attack is more difficult than the untargeted attack. As $\gamma$ decreases, the number of elements with flipped state is reduced, leading to smaller perturbation. Whereas, the impact of $\gamma$ on the attack success rate depends heavily on the attack scenario and the dataset. For the easier untargeted attack, it seems that a slightly large $\gamma$ is already helpful. The attack success rate will be saturated close to 100\% even if at $\gamma=0.01$. For the targeted attack with higher difficulty, it seems that there exists an obvious peak success rate on these datasets where the $\gamma$ value equals 0.05. The results are reasonable since the impact of $\gamma$ is two-fold: i) a too large $\gamma$ will result in a large perturbation amplitude and might cause a non-convergent attack; ii) a too small $\gamma$ cannot move the model out of the region with gradient vanishing.

\begin{figure}[!htbp]
\centering
\includegraphics[width=0.46\textwidth]{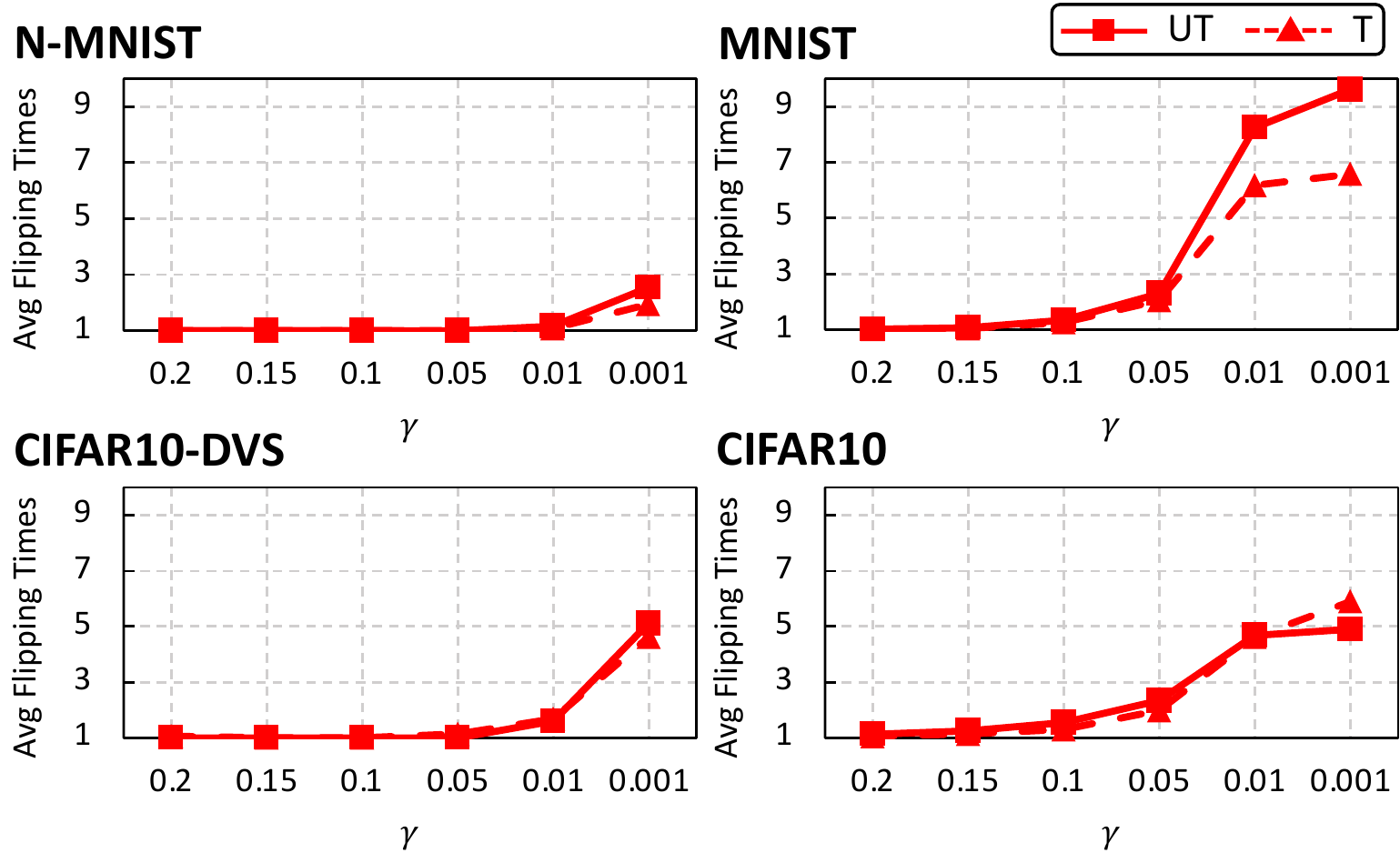}
\caption{Flipping times with different $\gamma$ settings in RSF. ``T'' and ``UT'', refer to targeted attack and untargeted attack, respectively. A smaller $\gamma$ increases the flipping times since the perturbation is not strong enough to push the model out of the gradient vanishing region, which needs more spike flipping.}
\label{fig:gt_result2}
\vspace{-10pt}
\end{figure}

 \begin{figure*}[!htbp]
\centering
\includegraphics[width=0.98\textwidth]{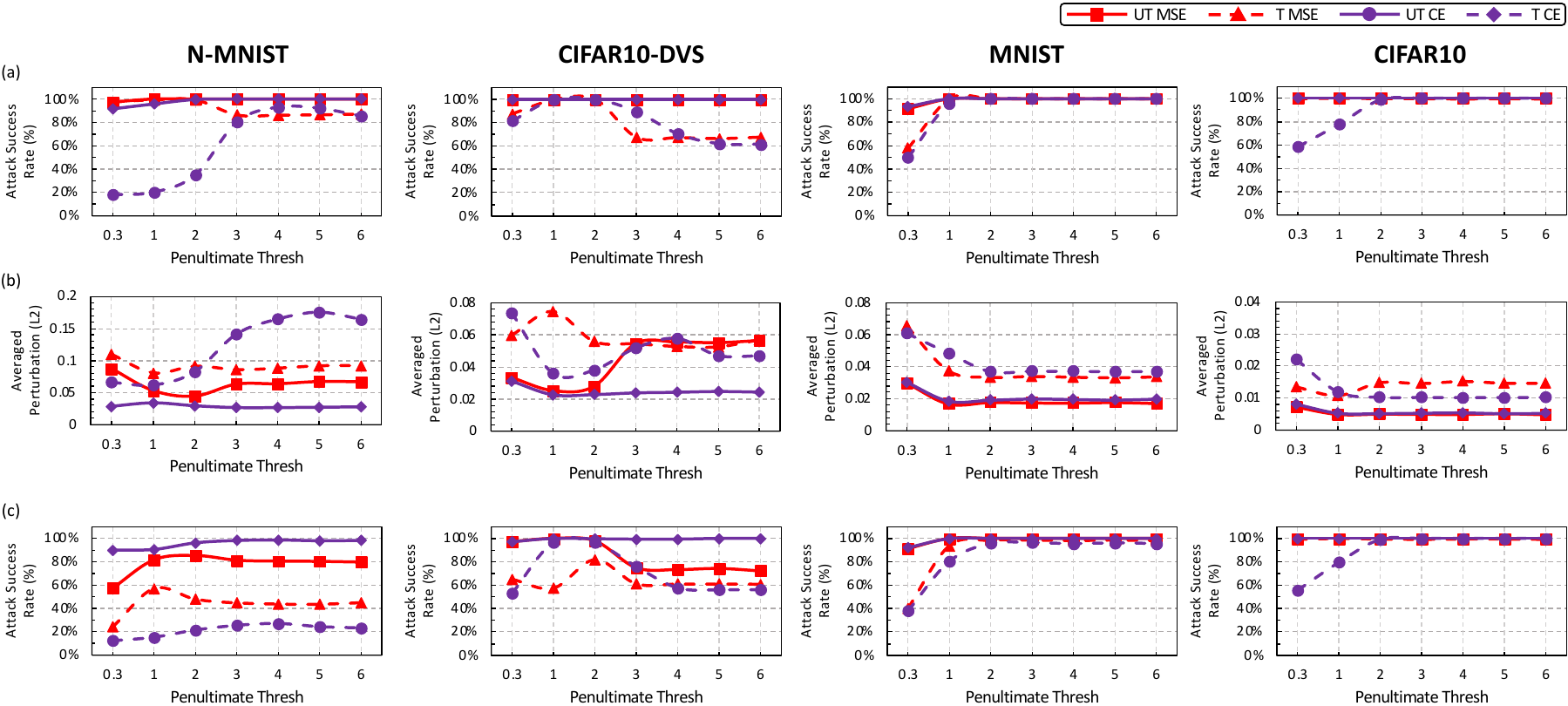}
\caption{Attack effectiveness with different firing threshold: (a) attack success rate without strict $\epsilon$ bound; (b) average perturbation without strict $\epsilon$ bound; (c) attack success rate under strict perturbation bound ($\epsilon=0.08$). ``T'' and ``UT'', refer to targeted attack and untargeted attack, respectively. In most cases, we can achieve a high attack success rate and acceptable perturbation with a slightly larger firing threshold at the penultimate layer, even if with strict perturbation bound ($\epsilon=0.08$).
} 
\label{fig:res:thre}
\end{figure*}

We also record the number of flipping times under different $\gamma$ setting, as shown in Figure \ref{fig:gt_result2}. Here the ``flipping times'' means the number of iterations during the attack process where the gradient vanishing occurs and the spike flipping is needed. We report the average value across different input examples. When $\gamma$ is large, the number of flipping times can be only one since the perturbation is large enough to push the model out of the gradient vanishing region. As $\gamma$ becomes smaller, the required number of flipping times becomes larger. In order to balance the attack success rate (see Figure \ref{fig:gt_result1}) and the flipping time (see Figure \ref{fig:gt_result2}), we finally recommend the setting of $\gamma=0.05$ in RSF on the datasets we tested. In real-world applications, the ideal setting should be explored again according to actual needs.

\subsection{Influence of Loss Function and Firing Threshold} \label{sec:exp:loss_threshold}

Additionally, we evaluate the influence of different training loss function on the attack success rate. The comparison is summarized in Table \ref{tab:res:loss}. Here the G2S converter and RSF are switched on. The model trained by CE loss leads to a lower attack success rate compared to the one trained by MSE loss, and the gap is especially large in the targeted attack scenario. As explained in Section \ref{sec:loss_thre}, this reflects the ``trap'' region of the models trained by CE loss due to the the increasing spike activities in the penultimate layer during attack.


  

  

           
          



\begin{table}[!htbp]
\caption{Impact of the training loss function on the attack success rate (without firing threshold optimization). ``T'' and ``UT'', refer to targeted attack and untargeted attack, respectively.}
\vspace{5pt}
\label{tab:res:loss}
\centering
\renewcommand\arraystretch{1.3}
\resizebox{0.43\textwidth}{!}{
\begin{tabular}{  c | c  c | c  c  } 

  \hline
  \hline
  
  & \multicolumn{2}{c|}{MSE Loss} & \multicolumn{2}{c}{CE Loss} \\

  Dataset & UT & T & UT & T \\
  \hline
  
  N-MNIST & 97.38\% & 99.44\%
          & 90.12\% & 16.78\% \\

 CIFAR10-DVS & 100\% & 86.35\%
             & 100\% & 82.95\% \\
           
 MNIST & 91.31\% & 55.33\% 
       & 93.16\% & 47.81\% \\
          
 CIFAR10 & 98.68\% & 99.72\%
       & 98.48\%  & 40.51\% \\

\hline
\hline

\end{tabular}}
\end{table}

To improve the attack effectiveness, we increase the firing threshold of the penultimate layer during attack to sparsify the spiking activities. Note that we only modify the penultimate layer’s firing threshold in the forward pass during the generation of adversarial examples. We use the original model without changing the threshold during attack with the generated adversarial examples, thus do not affect the network accuracy. The experimental results are provided in Figure \ref{fig:res:thre}. For untargeted attack, the increase of the firing threshold can improve the attack success rate to almost 100\% on all datasets. For targeted attack, the cases present different behaviors. Specifically, on image datasets (i.e. MNIST and CIFAR10), the attack success rate can be quickly improved and remained at about 100\%; while on spiking datasets (i.e. N-MNIST and CIFAR10-DVS), the attack success rate initially goes higher and then decreases, in other words, there exists a best threshold setting. This might be due to the sparse-event nature of the neuromorphic datasets, on which the number of spikes injected into the last layer will be decreased severely if the firing threshold becomes large enough, leading to a fixed loss value and thus a degraded attack success rate. Moreover, from the perturbation distribution, it can be seen that the increase of the firing threshold does not introduce much extra perturbation in most cases. All the above results indicate that appropriately increasing the firing threshold of the penultimate layer is able to improve the attack effectiveness significantly without enlarging the perturbation. 

In Figure \ref{fig:res:thre}(a), we do not strictly bound $\epsilon$, in order to avoid disturbing the analysis of the firing threshold. The average perturbation magnitude values are shown in Figure \ref{fig:res:thre}(b), which are relatively small (within 0.1 in most cases). We further analyze the attack success rate under the limitation of strict perturbation bounds. Specifically, during attack iterations, if the average perturbation per point is greater than a pre-defined value $\epsilon$, the attack is considered as a failure. As shown in Figure \ref{fig:res:thre}(c),  results show that our attack method can still achieve a considerable attack success rate with $\epsilon=0.08$ when compared to the results in Figure \ref{fig:res:thre}(a) for most cases. While there is a degradation for targeted attack over the N-MNIST dataset, which may be caused by the high sparsity of the spike inputs in that dataset. Overall, we achieve high attack success rate (more than 96\%) on image-based inputs and initiate the attack demonstration on spike-based inputs. We visualize several successful adversarial examples in the Appendix  (see Figure \ref{fig:res:adversarial_examples}).

\subsection{Effectiveness Comparison with Existing SNN Attack}

As discussed in Section \ref{sec:cha:comparison}, our attack is quite different from previous work using trial-and-error input perturbation \cite{marchisio2019snn,bagheri2018adversarial} or SNN/ANN model conversion \cite{sharmin2019comprehensive}. Beyond the methodology difference, here we coarsely discuss the attack effectiveness. Due to the high complexity of the trial-and-error manner, the testing dataset is quite small (e.g. USPS dataset \cite{marchisio2019snn}) or even with only one single example \cite{bagheri2018adversarial}. In contrast, we demonstrate the effective adversarial attack on much larger datasets including MNIST, CIFAR10, N-MNIST, and CIFAR10-DVS. For the SNN/ANN model conversion method \cite{sharmin2019comprehensive}, the authors show results on the CIFAR10 dataset. In that work, the authors used the accuracy loss of the model, which is caused by substituting the original normal inputs with the generated adversarial examples, for the evaluation of the attack effectiveness. We compare our attack results with theirs (inferred from the figure data in \cite{sharmin2019comprehensive}) on CIFAR10 under different $\epsilon$ configurations, as shown in Table \ref{tab:res:roy}. It can be seen that our attack method can incur more model accuracy loss in most cases, which indicates our better attack effectiveness.

\begin{table}[!htbp]
\caption{Comparison of the accuracy loss between our work and prior work \cite{sharmin2019comprehensive} under different perturbation bounds.}
\vspace{3pt}
\label{tab:res:roy}
\centering
\renewcommand\arraystretch{1.3}
\resizebox{0.46\textwidth}{!}{
\begin{tabular}{  c | c | c | c | c  } 

  \hline
  \hline

  $\epsilon$ & 8/255 & 16/255 & 32/255 & 64/255 \\
  \hline
   Untargeted \cite{sharmin2019comprehensive} & 37.50\% & 62.50\% & 75.00\% & 77.00\% \\
   \hline
  Untargeted (ours) & 50.47\% & 72.46\% & 76.67\% & 76.86\% \\
   \hline
  Targeted \cite{sharmin2019comprehensive} & 20.00\% & 37.50\% & 52.50\% & 63.00\% \\
  \hline
  Targeted (ours) & 19.16\% & 42.36\% & 65.58\% & 71.48\% \\

\hline
\hline

\end{tabular}}
\end{table}

\subsection{Effectiveness Comparison with ANN Attack}

In this subsection, we further compare the attack effectiveness between SNNs and ANNs on image-based datasets, i.e. MNIST and CIFAR10. For ANN models, we use the same network structure as SNN models given in Table \ref{tab:net_structure}. The training loss function is CE here. We test two attack scenarios: independent attack and cross attack. For the independent attack, the ANN models are attacked using the BIM method in Equation (\ref{equ:bim}); while the SNN models are attacked using the proposed methodology. Note that the firing threshold of the the penultimate layer of SNN models during attack is set to 2 in this subsection as suggested by Figure \ref{fig:res:thre}. For the cross attack, we use the adversarial examples generated by attacking the SNN models to mislead the ANN models, or vice versa.

\begin{figure}[!htbp]
\centering
\includegraphics[width=0.45\textwidth]{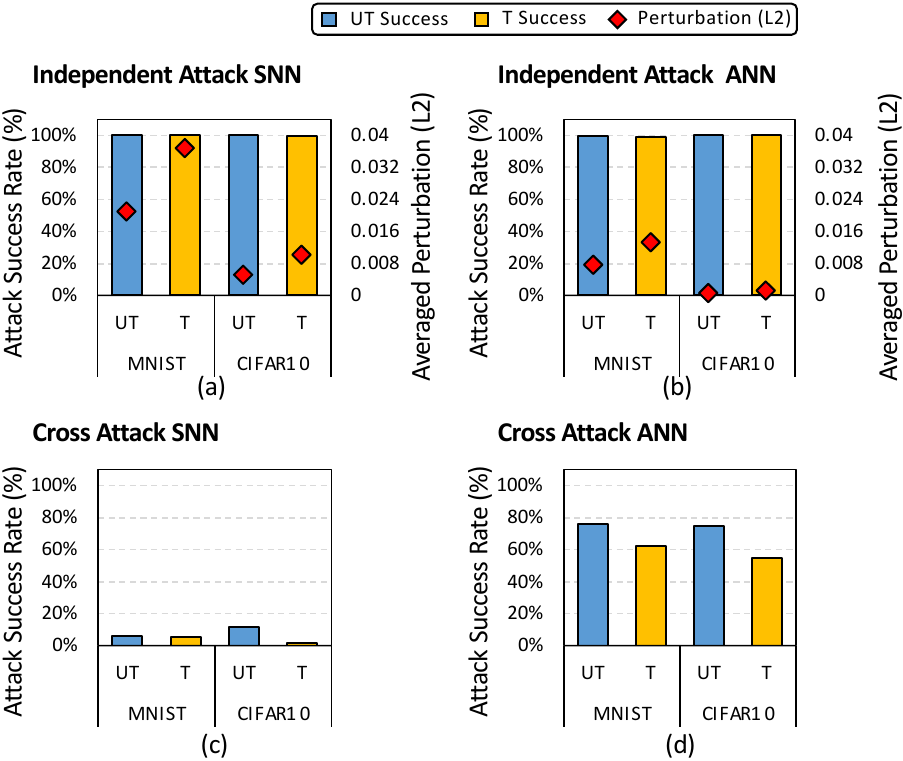}
\caption{Attack success rate comparison between ANNs and SNNs. ``T'' and ``UT'', refer to targeted attack and untargeted attack, respectively. Attacking SNNs requires larger perturbation than attacking ANNs and the adversarial examples generated by attacking the ANN models fail to attack the SNN models, which empirically reflect the higher robustness of SNNs.}
\label{fig:res:compare_ANN}
\end{figure}

From Figure \ref{fig:res:compare_ANN}(a)-(b), we can easily observe that all attack success rates are quite high in the independent attack scenario. While, attacking the SNN models requires larger perturbation than attacking the ANN models in the above experiment, which empirically reflects the higher robustness of the SNN models. From the results of the cross attack in Figure \ref{fig:res:compare_ANN}(c)-(d), we find that using the adversarial examples generated by attacking ANN models to fool the SNN models is very difficult, with only $<$12\% success rate. This further helps conclude that attacking an SNN model is harder than attacking an ANN model with the same network structure. The robustness of SNN may be jointly achieved by two factors: (1) the binarization of neuronal activities which is similar to the robustness of quantized ANNs against adversarial attack; (2) the leakage-and-fire mechanism which can naturally filter out the non-strong input noise. \cite{panda2019discretization, khalil2018combinatorial, galloway2017attacking}

\subsection{Other Attack Methods and Datasets}

In this subsection, we validate the effectiveness of the proposed attack methodology using more experiments with advanced attack methods (e.g., CWL2~\cite{carlini2017towards}) and dynamic datasets (e.g., Gesture-DVS~\cite{amir2017low}).


CWL2 is an advanced adversarial attack method that is widely applied in ANN attack, which integrates a regularization item to restrict the magnitude of the perturbation. We tailor Algorithm~\ref{alg:overall_attack} to perform CWL2 attack against SNN models over image-based inputs. The adversarial example generation follows
\begin{equation}
    xi'_{k+1} = xi'_k + \delta i_k - c \times \triangledown_{xi'_k} \lVert xi'_k - xi'_0 \rVert_2^2,
\end{equation}
where $xi'_0$ and $xi'_k$ represent the original input and the adversarial example generated at the $k$th attack iteration. $c$ is a parameter that determines the impact of regularization item. A larger $c$ indicates smaller perturbation at the cost of possibly lower attack success rate. The CWL2 attack would degrade to the classic BIM attack when $c=0$.

\begin{figure}[!htbp]
\centering
\includegraphics[width=0.48\textwidth]{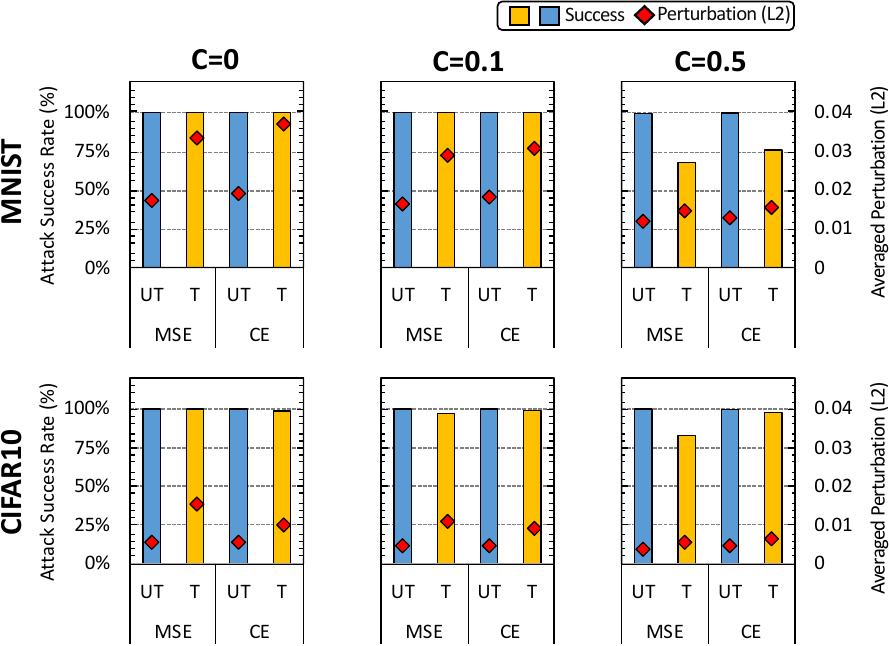}
\caption{Attack effectiveness with CWL2 method on MNIST and CIFAR10 under different settings of $c$. A larger $c$ indicates a smaller perturbation but possibly compromising the attack success rate. }
\label{fig:res:cw}
\end{figure}

We tested the tailored SNN-oriented CWL2 attack on MNIST and CIFAR10 datasets with different configurations of $c$. As illustrated in Figure \ref{fig:res:cw}, a slight increase of $c$ ($c=0.1$) helps reduce the perturbation magnitude without sacrificing the attack success rate, compared to the results at ($c=0$). However, when $c$ is too large ($c=0.5$), the attack success rate decreases. For example, the targeted attack success rate on MNIST dataset is reduced by up to 32.45\% when $c=0.5$.

\begin{figure}[!htbp]
\centering
\includegraphics[width=0.48\textwidth]{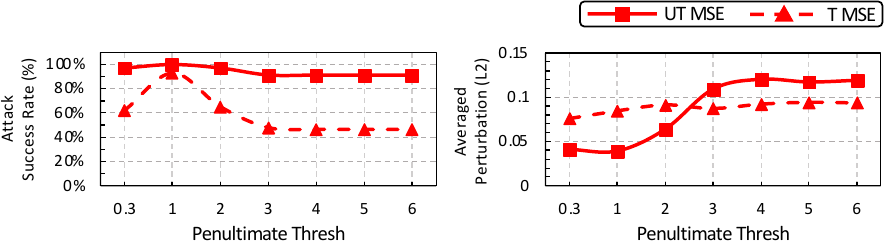}
\caption{Attack effectiveness on the Gesture-DVS dataset. Our methodology can achieve high attack success rate and acceptable perturbation with an appropriate setting of the firing threshold at the penultimate layer even in this scenario.}
\label{fig:res:gesture}
\end{figure}

In addition, we also applied our attack method on Gesture-DVS. The model configurations have been shown in Table~\ref{tab:para_acc}. We only test the cases with MSE training loss function for simplicity, and the attack results are shown in Figure \ref{fig:res:gesture}. Our methodology can still achieve a high attack success rate with acceptable perturbation even on this dynamic dataset. The trend of attack success rate variation under different penultimate layer threshold setting is similar to that on other spike-based datasets we have tested earlier.

%% file: text/Conclusion.tex
\section{Conclusion and Discussion}\label{sec:Conclusion}

SNNs have attracted broad attention and have been widely deployed in neuromorphic devices due to the importance for brain-inspired computing. Naturally, the security problem of SNNs should be considered. In this work, we select the adversarial attack against the SNNs trained by BPTT-like supervised learning as a starting point. First, we identify the challenges in attacking an SNN model, i.e. the incompatibility between the spiking inputs and the continuous gradients, and the gradient vanishing problem. Second, we design a gradient-to-spike (G2S) converter with probabilistic sampling, sign extraction, and overflow-aware transformation, and a restricted spike flipper (RSF) with element selection and gradient construction to address the mentioned two challenges, respectively. Our methodology can control the perturbation amplitude well and is applicable to both spiking and image data formats. Interestingly, we find that there is a ``trap'' region in SNN models trained by CE loss, which can be overcome by adjusting the firing threshold of the penultimate layer. We conduct extensive experiments on various datasets including MNIST, CIFAR10, N-MNIST, and CIFAR10-DVS and show 99\%+ attack success rate in most cases, which is the best result on SNN attack. The in-depth analysis on the influence of G2S converter, RSF, loss function, and firing threshold are also provided. Furthermore, we compare the attack of SNNs and ANNs and reveal the robustness of SNNs against adversarial attack. Our findings are helpful to understand the SNN attack and can stimulate more research on the security of neuromorphic computing.

For future work, we recommend several interesting topics. Although we only study the white-box adversarial attack to avoid shifting the focus of presenting our methodology, the black-box adversarial attack should be investigated because it is more practical. Fortunately, the proposed methods in this work can be transferred into the black-box attack scenario. Second, we only analyze the influence of loss function and firing threshold due to the page limit. It still remains an open question that whether other factors can affect the attack effectiveness, such as the gradient approximation form of the firing activities, the time window length for rate coding or the coding scheme itself, the network structure, and other solutions that can substitute G2S and RSF. Third, the attack against physical neuromorphic devices rather than just theoretical models is more attractive. At last, compared to the attack methods, the defense techniques are highly expected for the construction of large-scale neuromorphic systems.

%% file: text/biography.tex
\vspace{-30pt}

\begin{IEEEbiography}[{\includegraphics[width=1in,height=1.25in,clip,keepaspectratio]{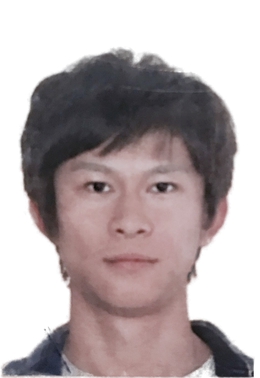}}] {Ling Liang} received the B.E. degree from Beijing University of Posts and Telecommunications, Beijing, China in 2015, and M.S. degree from University of Southern California, CA, USA in 2017. He is currently pursuing the Ph.D. degree at Department of Electrical and Computer Engineering,  University of California, Santa Barbara, CA, USA. His current research interests include machine learning security, tensor computing, and computer architecture.
\end{IEEEbiography}
\vspace{-30pt}

\begin{IEEEbiography}[{\includegraphics[width=1in,height=1.25in,clip,keepaspectratio]{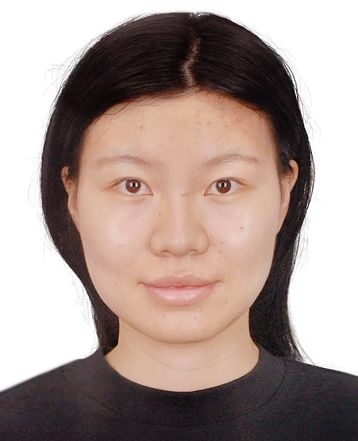}}] {Xing Hu} received the B.S. degree from Huazhong University of Science and Technology, Wuhan, China, and Ph.D. degree from University of Chinese Academy of Sciences, Beijing, China in 2009 and 2014, respectively. She is currently a Postdoctoral Fellow at the Department of Electrical and Computer Engineering, University of California, Santa Barbara, CA, USA. Her current research interests include emerging memory system, domain-specific hardware, and machine learning security.
\end{IEEEbiography}
\vspace{-30pt}

\begin{IEEEbiography}[{\includegraphics[width=1in, height=1.25in, clip, keepaspectratio]{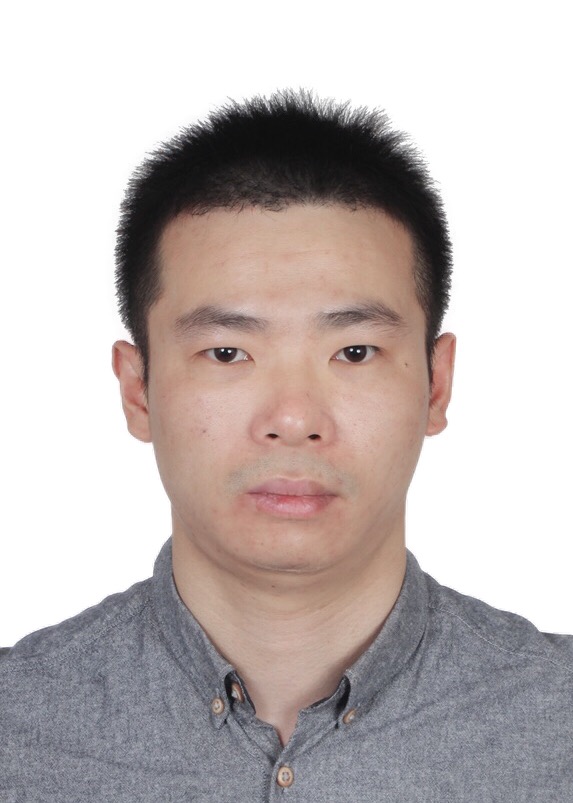}}] {Lei Deng} received the B.E. degree from University of Science and Technology of China, Hefei, China in 2012, and the Ph.D. degree from Tsinghua University, Beijing, China in 2017. He is currently a Postdoctoral Fellow at the Department of Electrical and Computer Engineering, University of California, Santa Barbara, CA, USA. His research interests span the areas of brain-inspired computing, machine learning, neuromorphic chip, computer architecture, tensor analysis, and complex networks. Dr. Deng has authored or co-authored over 50 refereed publications. He was a PC member for \emph{International Symposium on Neural Networks (ISNN)} 2019. He currently serves as a Guest Associate Editor for \emph{Frontiers in Neuroscience} and \emph{Frontiers in Computational Neuroscience}, and a reviewer for a number of journals and conferences. He was a recipient of MIT Technology Review Innovators Under 35 China 2019. 
\end{IEEEbiography}
\vspace{-30pt}

\begin{IEEEbiography}[{\includegraphics[width=1in, height=1.25in, clip, keepaspectratio]{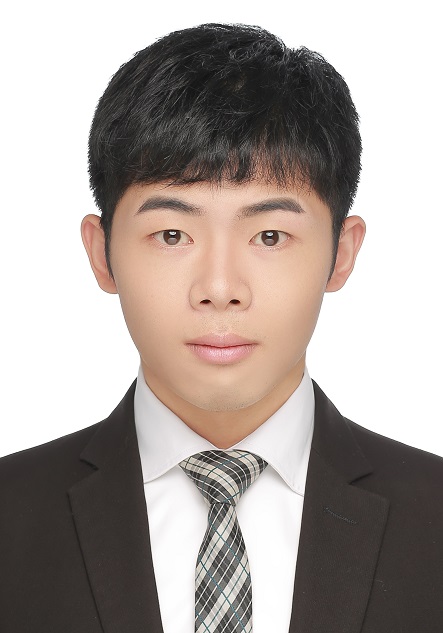}}] {Yujie Wu} received the B.E. degree in Mathematics and Statistics from Lanzhou University, Lanzhou, China in 2016. He is currently pursuing the Ph.D. degree at the  Center  for  Brain Inspired  Computing  Research  (CBICR), Department of Precision Instrument, Tsinghua University, Beijing, China. His current research interests include spiking neural networks, neuromorphic device, and brain-inspired computing.
\end{IEEEbiography}
\vspace{-30pt}

\begin{IEEEbiography}[{\includegraphics[width=1in,height=1.25in,clip,keepaspectratio]{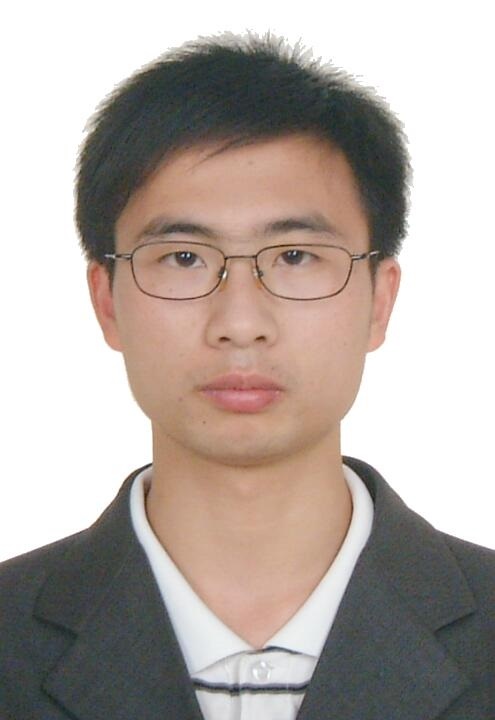}}]{Guoqi Li}  received the B.E. degree from the Xi’an University of Technology, Xi’an, China in 2004, the M.E. degree from Xi’an Jiaotong University, Xi’an, China in 2007, and the Ph.D. degree from Nanyang Technological University, Singapore, in 2011. He was a Scientist with Data Storage Institute and the Institute of High Performance Computing, Agency for Science, Technology and Research (ASTAR), Singapore from 2011 to 2014. He is currently an Associate Professor with Center for Brain Inspired Computing Research (CBICR), Tsinghua University, Beijing, China. His current research interests include machine learning, brain-inspired computing, neuromorphic chip, complex systems and system identification. Dr. Li is an Editorial-Board Member for \emph{Control and Decision} and a Guest Associate Editor for \emph{Frontiers in Neuroscience, Neuromorphic Engineering}. He was the recipient of the 2018 First Class Prize in Science and Technology of the Chinese Institute of Command and Control, Best Paper Awards (\emph{EAIS} 2012 and \emph{NVMTS} 2015), and the 2018 Excellent Young Talent Award of Beijing Natural Science Foundation. 
\end{IEEEbiography}
\vspace{-30pt}

\begin{IEEEbiography}[{\includegraphics[width=1in, height=1.25in, clip, keepaspectratio]{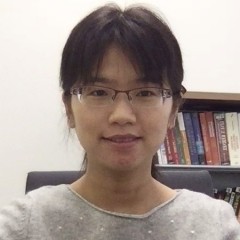}}] {Yufei Ding} received her B.S. degree in Physics from University of Science and Technology of China, Hefei, China in 2009, M.S. degree from The College of William and Mary, VA, USA in 2011, and the Ph.D. degree in Computer Science from North Carolina State University, NC, USA in 2017. She joined the Department of Computer Science, University of California, Santa Barbara as an Assistant Professor since 2017. Her research interest resides at the intersection of Compiler Technology and (Big) Data Analytics, with a focus on enabling High-Level Program Optimizations for data analytics and other data-intensive applications. She was the receipt of NCSU Computer Science Outstanding Research Award in 2016 and Computer Science Outstanding Dissertation Award in 2018.
\end{IEEEbiography}
\vspace{-35pt}

\begin{IEEEbiography}[{\includegraphics[width=1in, height=1.25in, clip, keepaspectratio]{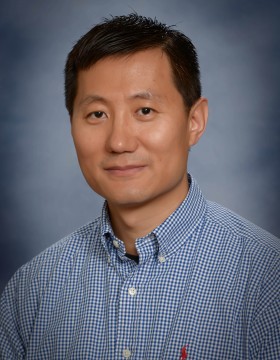}}] {Peng Li} received the Ph.D. degree in electrical and computer engineering from Carnegie Mellon University, Pittsburgh, PA, USA in 2003. He was a Professor with Department of Electrical and Computer Engineering, Texas A\&M University, College Station, TX, USA from 2004 to 2019. He is presently a Professor with Department of Electrical and Computer Engineering, University of California, Santa Barbara, CA, USA. His research interests include integrated circuits and systems, computer-aided design, brain-inspired computing, and computational brain modeling.

His work has been recognized by various distinctions including the ICCAD Ten Year Retrospective Most Influential Paper Award, four IEEE/ACM Design Automation Conference Best Paper Awards, the IEEE/ACM William J. McCalla ICCAD Best Paper Award, the ISCAS Honorary Mention Best Paper Award from the Neural Systems and Applications Technical Committee of IEEE Circuits and Systems Society, the US National Science Foundation CAREER Award, two Inventor Recognition Awards from Microelectronics Advanced Research Corporation, two Semiconductor Research Corporation Inventor Recognition Awards, the William and Montine P. Head Fellow Award and TEES Fellow Award from the College of Engineering, Texas A\&M University. He was an Associate Editor for \emph{IEEE Transactions on Computer-Aided Design of Integrated Circuits and Systems} from 2008 to 2013 and \emph{IEEE Transactions on Circuits and Systems-II: Express Briefs} from 2008 to 2016, and he is currently a Guest Associate Editor for \emph{Frontiers in Neuroscience}. He was the Vice President for Technical Activities of IEEE Council on Electronic Design Automation from 2016 to 2017.
\end{IEEEbiography}
\vspace{-35pt}

\begin{IEEEbiography}[{\includegraphics[width=1in,height=1.25in,clip,keepaspectratio]{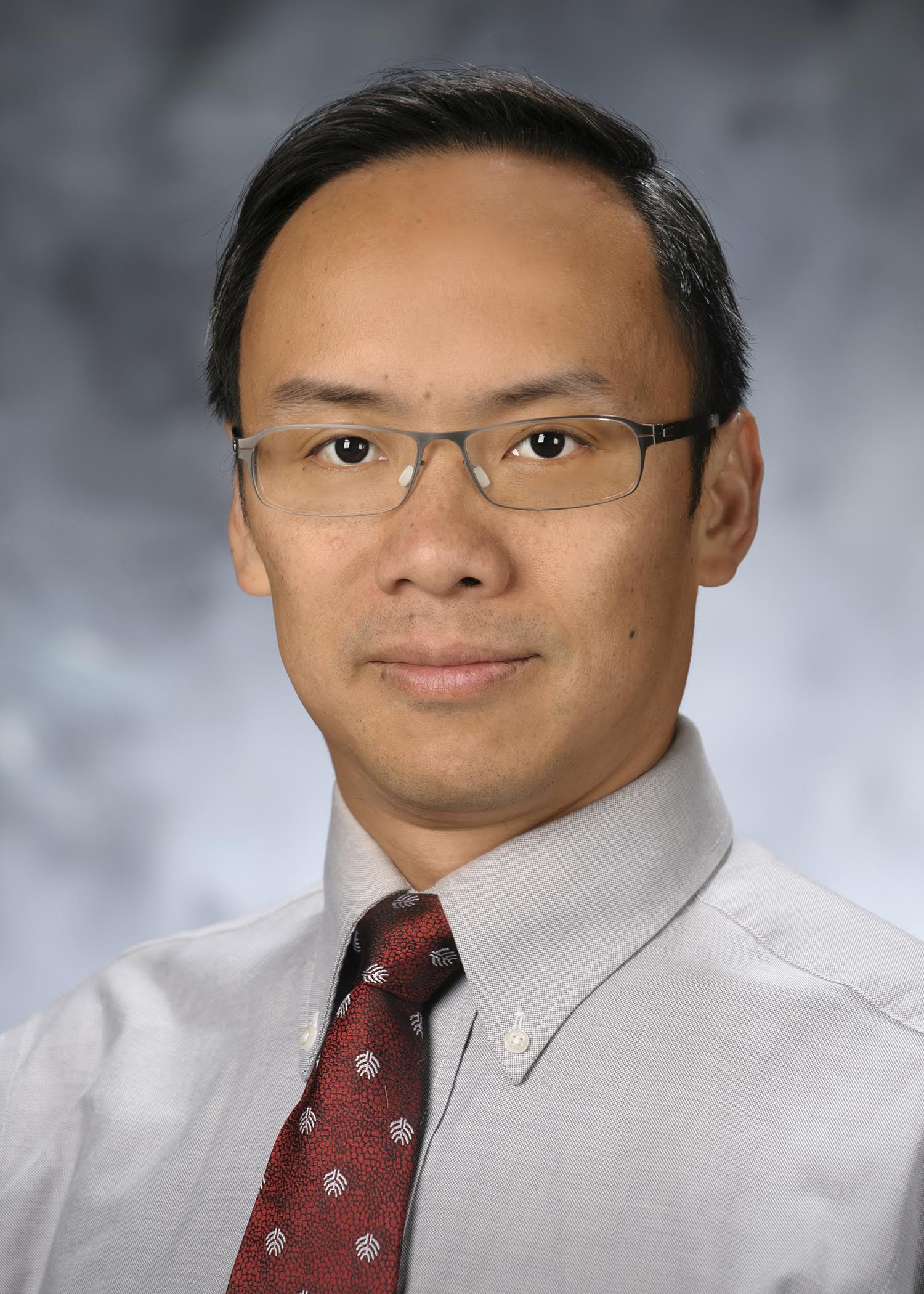}}]{Yuan Xie}  received the B.S. degree in Electronic Engineering from Tsinghua University, Beijing, China in 1997, and M.S. and Ph.D. degrees in Electrical Engineering from Princeton University, NJ, USA in 1999 and 2002, respectively. He was an Advisory Engineer with IBM Microelectronic Division, Burlington, NJ, USA from 2002 to 2003. He was a Full Professor with Pennsylvania State University, PA, USA from 2003 to 2014. He was a Visiting Researcher with Interuniversity Microelectronics Centre (IMEC), Leuven, Belgium from 2005 to 2007 and in 2010. He was a Senior Manager and Principal Researcher with AMD Research China Lab, Beijing, China from 2012 to 2013. He is currently a Professor with the Department of Electrical and Computer Engineering, University of California at Santa Barbara, CA, USA. His interests include VLSI design, Electronics Design Automation (EDA), computer architecture, and embedded systems. 

Dr. Xie is an expert in computer architecture who has been inducted to \emph{ISCA}/\emph{MICRO}/\emph{HPCA} Hall of Fame and IEEE/AAAS/ACM Fellow. He was a recipient of Best Paper Awards (\emph{HPCA} 2015, \emph{ICCAD} 2014, \emph{GLSVLSI} 2014, \emph{ISVLSI} 2012, \emph{ISLPED} 2011, \emph{ASPDAC} 2008, \emph{ASICON} 2001) and Best Paper Nominations (\emph{ASPDAC} 2014, \emph{MICRO} 2013, \emph{DATE} 2013, \emph{ASPDAC} 2010-2009, \emph{ICCAD} 2006), the 2016 IEEE Micro Top Picks Award, the 2008 IBM Faculty Award, and the 2006 NSF CAREER Award. He served as the TPC Chair for \emph{ICCAD} 2019, \emph{HPCA} 2018, \emph{ASPDAC} 2013, \emph{ISLPED} 2013, and \emph{MPSOC} 2011, a committee member in IEEE Design Automation Technical Committee (DATC), the Editor-in-Chief for \emph{ACM Journal on Emerging Technologies in Computing Systems}, and an Associate Editor for \emph{ACM Transactions on Design Automations for Electronics Systems}, \emph{IEEE Transactions on Computers}, \emph{IEEE Transactions on Computer-Aided Design of Integrated Circuits and Systems}, \emph{IEEE Transactions on VLSI, IEEE Design and Test of Computers}, and \emph{IET Computers and Design Techniques}. Through extensive collaboration with industry partners (e.g. AMD, HP, Honda, IBM, Intel, Google, Samsung, IMEC, Qualcomm, Alibaba, Seagate, Toyota, etc.), he has helped the transition of research ideas to industry. 
\end{IEEEbiography}

%% file: text/Appendix.tex
\newpage
\section{Appendix}

 \begin{figure*}[!htbp]
\centering
\includegraphics[width=0.98\textwidth]{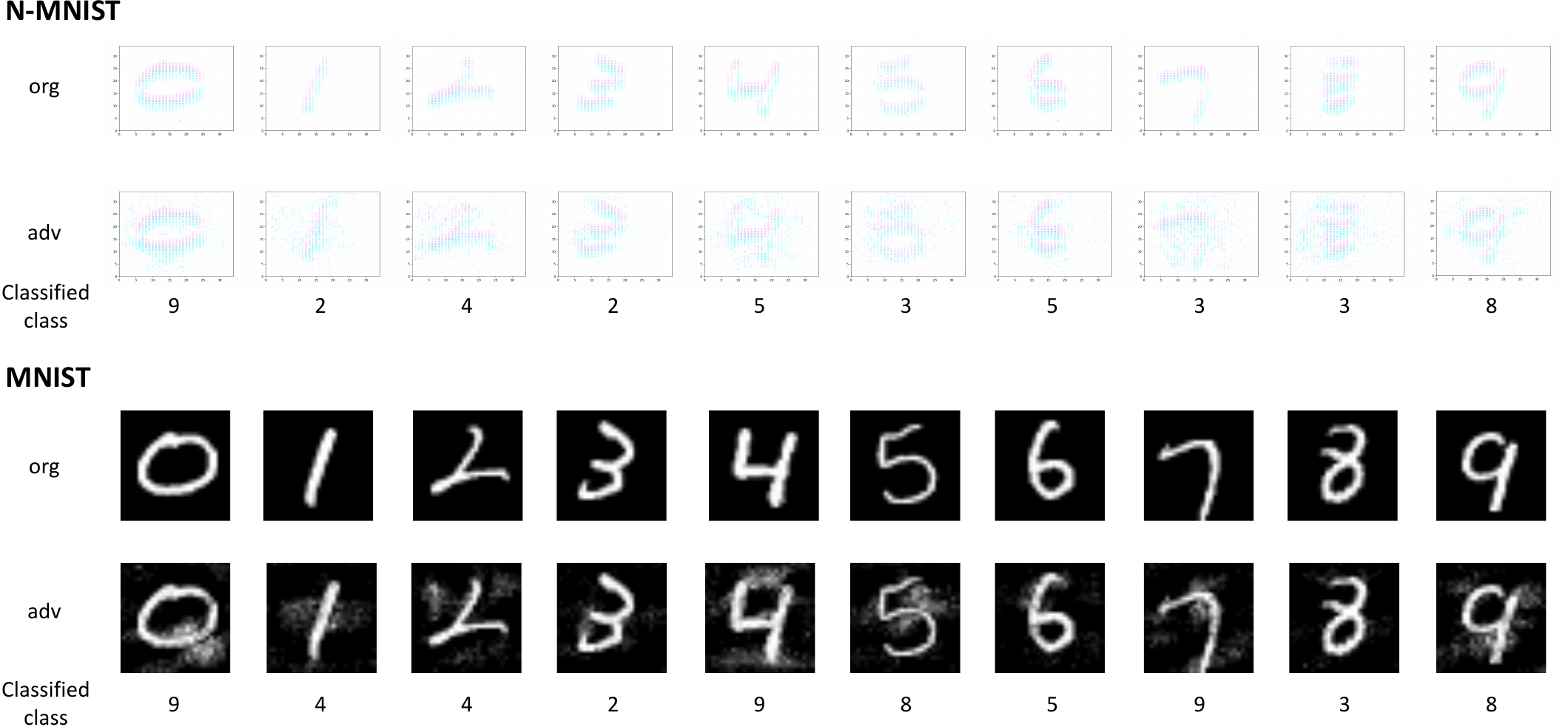}
\caption{Example of original samples and successful adversarial samples generated through our attack method in N-MNIST and MNIST dataset under untargeted attack. The penultimate layer threshold during attack is set to 2. In N-MNIST dataset, we squeeze the sample along time steps.}
\label{fig:res:adversarial_examples}
\end{figure*}